\pdfoutput=1
\documentclass[letterpaper]{article}
\usepackage[preprint]{aaai2027}
\usepackage[hyphens]{url}  
\usepackage{graphicx}      
\usepackage{natbib}        
\usepackage{caption}       
\usepackage{amsmath,amssymb}
\usepackage{booktabs}
\usepackage{tabularx}
\usepackage[table]{xcolor}
\definecolor{StageBlue}{HTML}{4F83CC}
\definecolor{StageGreen}{HTML}{4A9A79}
\definecolor{StageOrange}{HTML}{E6A04B}
\definecolor{StageRed}{HTML}{C94A5A}
\definecolor{StageGray}{HTML}{6B7785}

\title{What Does Post-Training Change in Multilingual Reasoning?}
\author{
    Hongyang Li\textsuperscript{\rm 1},
    Xiao Li\textsuperscript{\rm 2},
    Caesar Wu\textsuperscript{\rm 1},
    Gr\'egoire Danoy\textsuperscript{\rm 1},
    Pascal Bouvry\textsuperscript{\rm 1}
}
\affiliations{}
\newcommand{\affilnote}{\begingroup\renewcommand{\thefootnote}{}\footnotetext{\raggedright
$^{1}$University of Luxembourg\par
\hspace*{0.6em}\texttt{\{hongyang.li,\,caesar.wu,\,gregoire.danoy,}\par
\hspace*{0.6em}\texttt{\phantom{\{}pascal.bouvry\}@uni.lu}\par
$^{2}$Seafill Open-Source Community, \texttt{xiao.li@seafill.com}}\endgroup}

\begin{document}
\maketitle
\affilnote
\raggedbottom

\begin{abstract}
Open-source reasoning models provide unequal access to reasoning capability across languages. When a model can solve a problem but cannot deliver a complete solution in the user’s language, language becomes an access barrier rather than merely a source of performance variation. We audit Qwen3 checkpoints on competition-mathematics tasks in eleven languages. Across the ten non-English languages, only 15.4--17.9\% of problems receive a correct, terminating solution with visible reasoning in the requested language in any of 16 samples, compared with 92.9\% in English. To identify the source of this disparity, we evaluate thirteen endpoints from one model family, spanning released checkpoints, multilingual supervised fine-tuning (SFT) at two scales, controlled SFT ablations, and three reinforcement-learning (RL) reward formulations. We jointly track correctness, language adherence, termination, and delivery efficiency. The dominant bottleneck shifts across post-training stages. Released models often reason in English. Multilingual SFT restores target-language reasoning, but accuracy declines across multilingual, English-only, and single-language SFT runs, showing that this cost is not specific to multilingual mixing; non-English reasoning traces additionally become prone to non-terminating loops. RL restores termination in both arms at no cost in accuracy, but only the arm whose reward includes a language term delivers: rewarding correctness alone returns the model to English. Together, these stages establish a constructive post-training path from English-pivoted capability to multilingual reasoning that is reliably delivered.
\end{abstract}

\section{Introduction}\label{sec:problem}

A reasoning system serves users across languages only if it returns a correct and complete solution in the language they request. Final-answer accuracy cannot establish this condition: it credits correct answers whose visible reasoning is written in English rather than the requested language, and it collapses non-termination into generic error. Nominal multilingual capability can therefore conceal unequal access to successful problem solving \citep{cocola2025,tokentax2025}, and unequal cost of obtaining it, since the same content occupies very different numbers of tokens depending on the language it is written in \citep{petrov2023tokenizer}. Prior work separately documents gaps in multilingual accuracy, target-language adherence, efficiency, and reliability. These findings are often grouped under a single ``multilingual reasoning gap,'' but they represent distinct failure modes: a model may fail to solve the problem, solve it through an English visible trace, or begin a target-language solution without completing it. Because these effects are typically observed on different systems or at isolated training stages, they do not reveal which failure mode is dominant at a given point in training, how that bottleneck changes after intervention, or what cost its repair introduces. We study these questions within one open-weight model family. We follow released Qwen3 checkpoints \citep{qwen3} through multilingual supervised fine-tuning (SFT), English-only and single-language controls, a second-scale replication, and three reinforcement-learning (RL) reward formulations. All endpoints are evaluated under the same target-language prompts, sampling protocol, and response budget. For each response, we inspect both the final answer and the visible reasoning trace. A response is delivered only if its final answer is correct ($C$), its visible solution follows the requested language ($L$), and its generation terminates ($T$):
\begin{equation}
J=C\wedge L\wedge T.
\label{eq:joint}
\end{equation}
Reporting these components alongside their conjunction reveals when an intervention repairs one failure mode by worsening another. We additionally measure delivery efficiency as the token cost of obtaining a joint success. The main contribution is a controlled, stage-by-stage analysis of how successive post-training procedures change multilingual reasoning within a single model family.

Specifically, we contribute (i) a delivery-centered evaluation that exposes up to 94 points of
correct-but-undelivered performance and jointly reports correctness, language adherence,
termination, and delivery efficiency; (ii) an SFT diagnosis that separates a general accuracy cost, common to multilingual, English-only, and single-language runs, from looping, which rises only in non-English traces and is the failure that reduces delivery efficiency; (iii) a comparison of RL reward design showing that rewarding correctness alone is not enough---once termination is restored the model returns to English---while a language bonus simply added to it is collected by wrong answers as well; gating the bonus on a correct answer closes that route; and (iv) a constructive post-training path that uses SFT
to repair adherence, RL to restore termination, and a correctness-gated reward to hold
accuracy while improving end-to-end multilingual delivery.

\begin{figure*}[t]
\centering
\includegraphics[width=\textwidth]{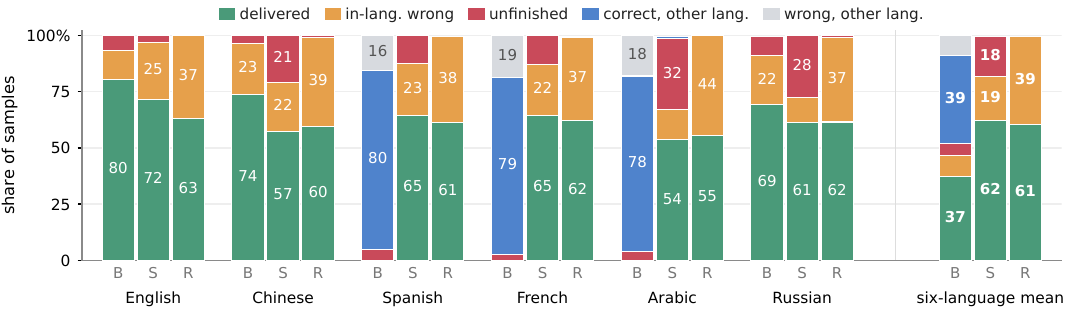}
\caption{\textbf{A constructive path to multilingual delivery (Qwen3-4B).}
B is the released checkpoint, S adds six-language multilingual SFT, and R is the final correctness-gated endpoint. Bars partition samples by their $J@1$ outcomes; non-terminating responses are counted as unfinished regardless of correctness or language. The released checkpoint often solves in the wrong language, SFT restores target-language delivery but introduces non-termination, and the final RL endpoint restores completion while retaining delivery. The rightmost group reports the six-language mean.}
\label{fig:summary}
\end{figure*}

Figure~\ref{fig:summary} summarizes the resulting path from the released checkpoint to the
correctness-gated endpoint.

\vspace{-6pt}
\section{Related Work}\label{sec:related}

Prior work documents English-centric internal representations
\citep{wendler2024llamas,schut2025think} and visible reasoning traces that often default to
high-resource languages irrespective of the input language \citep{languagematters2025}.
Work on long CoT directly contrasts English-pivoted and target-language reasoning traces
\citep{longcotlang2025}. Prior work
also attributes multilingual gaps to implicit translation at both ends of the pipeline: at
the output stage, where a task is solved and then rendered imperfectly in the target
language \citep{translationbarrier2025}, and at the input stage, where a non-English prompt
is not carried into the English-dominant reasoning trace \citep{multilingualgap2026}. A
further line shows that prompting models to reason in the user's language can reduce
answer accuracy \citep{whenmodelsreason2025}; suppressing within-trace language mixing can
also reduce accuracy in bilingual reasoning models \citep{languagemixing2025}. The PolyMath benchmark further reports
per-language accuracy, input--output language consistency, and reasoning length
\citep{polymath2025}. Related work also scores reasoning-language fidelity separately from
answer accuracy \citep{bridge2025}, while MTM-Bench decomposes semantic correctness,
target-language adherence, and joint success under crossed instruction, content, and
response languages \citep{mtmbench2026}. Language adherence and tokenization efficiency bear on equitable access to model capability
\citep{cocola2025,petrov2023tokenizer,tokentax2025}. Work on long chain-of-thought
training reports that limited long-chain-of-thought supervision degrades small models and
can impair subsequent RL \citep{throughvalley2025}, and that reasoning
models fall into repetition loops that consume the generation budget instead of terminating
\citep{loop2025}. Multilingual reinforcement learning couples
language behavior and task success through consistency-enhanced objectives and
language-adaptive guidance \citep{mthinker2026,lang2026}, including unsupervised
rewards that require only cross-lingual agreement of final answers
\citep{xlingconsistency2026}; closest to our setting, the reasoning language itself becomes
the training target \citep{reasonxl2026}. Taken together, these works establish the failure modes relevant to multilingual reasoning
but largely study them separately. We instead provide a controlled, stage-by-stage analysis
of one model family across released checkpoints, SFT, and RL. By holding the evaluation
fixed throughout, we identify which bottleneck dominates at each post-training stage and how
successive interventions change it.

\vspace{-6pt}
\section{Setup}\label{sec:method}

We evaluate thirteen endpoints from one Qwen3 family under a fixed protocol. Two are released checkpoints. Eight are SFT endpoints: six-language mixtures on Qwen3-4B and Qwen3-8B, an English-only Qwen3-4B model, and five Qwen3-4B specialists trained on one target language each. The remaining three are Qwen3-4B RL checkpoints from two arms trained on the same multilingual SFT checkpoint: a correctness-only control, and a target-language procedure run in two phases, an additive phase followed by a correctness-gated phase that continues it. All three are evaluated under the same protocol.

Every endpoint is evaluated on the same 70 competition-mathematics problems in every evaluation language, with 16 samples per problem, temperature 0.7, and a 24,576-token evaluation budget. The primary evaluation covers the six official UN languages: English, Chinese, Spanish, French, Arabic, and Russian; the five single-language specialists are evaluated only in their own target language and English (2,240 responses each). Each problem, instruction, and answer-format requirement is written in the requested language, and every visible reasoning trace is retained. For each response, $C$ denotes final-answer correctness, $L$ target-language adherence of the visible reasoning trace, and $T$ normal termination without reaching the token limit or entering a detected loop. Their response-level conjunction is $J$ (Equation~\ref{eq:joint}). We compute $J$ directly for each response rather than multiplying marginal rates, which would assume independence among correctness, adherence, and termination.

For target-language SFT, we begin with 45K English OpenR1 \citep{openr1} long-chain-of-thought mathematics traces and assign them to the six evaluation languages in round-robin order. English traces remain unchanged; the other five versions are translated by a locally deployed Qwen3-14B. A controlled comparison with Hunyuan-MT-7B \citep{hunyuanmt} (one of the most widely downloaded open-source translation models) finds Qwen3-14B better on every criterion, and we apply quality gates for answer preservation, reasoning structure, target language, and repetition. After filtering, the Qwen3-4B multilingual run uses 43,218 traces, with 6,976--7,214 examples in each translated language and 7,460 in English; the Qwen3-8B replication uses a separately constructed corpus of 56,307 traces.

Two Qwen3-4B controls help diagnose SFT effects. The English-only model uses 9,800 English traces and contains no target-language training data. Five specialists use 9k-10k traces from one target language each, sharing source problems with the multilingual corpus while differing in language composition. All models are trained with standard token-level SFT; complete corpus statistics and optimization recipes appear in Appendix~\ref{app:training}.

For target-language RL, we train with PPO \citep{ppo} on 16,754 DAPO-Math-17K \citep{dapo} prompts distributed round robin across the six languages. Each prompt, instruction, and answer-format suffix is written in the requested language, while the language-neutral gold answer remains unchanged. The control and the two phases of the target-language procedure use, respectively:
\begin{equation}
\begin{aligned}
R_{\mathrm{corr}}&=C_{\mathrm{tr}},\\[1pt]
R_{\mathrm{add}}\;&=0.7\,C_{\mathrm{tr}}+0.3\,\ell_{\mathrm{tr}},\\[1pt]
R_{\mathrm{gate}}&=C_{\mathrm{tr}}\,(0.7+0.3\,\ell_{\mathrm{tr}}),
\end{aligned}
\label{eq:rewards}
\end{equation}
In Equation~\ref{eq:rewards}, $C_{\mathrm{tr}}$ is training-time correctness and $\ell_{\mathrm{tr}}\in[0,1]$ is a continuous target-language score. $R_{\mathrm{corr}}$ is the control. The other two are the phases of one procedure: the additive phase shapes densely, awarding up to $0.3$ language credit to an incorrect response, and the gated phase withdraws that credit, assigning every incorrect response zero reward.

The control and the additive phase use a 16,384-token training cap and batch size 32. The gated phase continues from the additive checkpoint at step 200 with an 8,192-token cap and batch size 64. All runs use clipped PPO with a scalar terminal reward, GAE \citep{gae} with $\gamma=\lambda=1$, and the same soft overlong-response penalty \citep{dapo} (a 2,048-token buffer with factor 1.0). Appendix~\ref{app:training} reports the complete SFT and PPO recipes (Tables~\ref{tab:recipesft} and~\ref{tab:reciperl}).

\begin{figure*}[t]
\centering
\includegraphics[width=\textwidth]{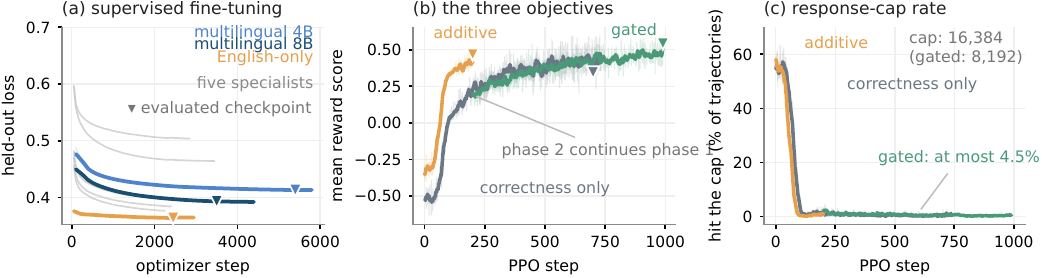}
\caption{\textbf{Training diagnostics for the eleven runs.} (a) Held-out loss for the eight
supervised runs, with the evaluated checkpoint marked on the three whose checkpoint the
paper reports. The Qwen3-4B and Qwen3-8B
multilingual runs are drawn in different blues; they use separately constructed corpora of
43,218 and 56,307 traces, so only the shape of each curve is comparable across scales.
(b) Mean PPO reward for the control and for the two phases of the target-language
procedure, all Qwen3-4B; the levels are not comparable because each optimizes its own
reward. (c) The share of
PPO trajectories reaching the response cap, which is 16,384 tokens for the correctness-only
and additive runs and 8,192 for the gated continuation. Pale lines are raw observations and
dark lines centered rolling means.}
\label{fig:training}
\end{figure*}

Table~\ref{tab:runs} lists every run with its initialization, data volume, batch size, and evaluated checkpoint; the supervised endpoints are read at epoch~2 for the Qwen3-4B multilingual run and epoch~1 for the Qwen3-8B replication, and the RL endpoints at steps 700, 200, and 1,000. Figure~\ref{fig:training} shows that each supervised run has reached a stable held-out loss at the checkpoint we evaluate and that each PPO run increases its own reward, so the comparisons that follow are not between a converged and an unconverged model.

Finally, we measure delivery efficiency as the number of joint successes produced per 1,000 generated tokens. For endpoint and language $\ell$,
\[
\mathrm{joint\_per\_1k}(\ell)
=
\frac{1000\times\#\{\text{responses with }J=1\}}
{\#\{\text{adjusted tokens}\}}.
\]
The denominator includes every attempt, including responses that never deliver. For a trace written in the requested language, tokens are divided by its language-specific encoding factor, measured on parallel prompts; English-pivoted traces receive no hypothetical target-language adjustment. Table~\ref{tab:tokeneff} reports the corresponding raw-token results and the efficiency conditioned on delivered responses; Appendix~\ref{app:additional} resolves both by language (Table~\ref{tab:efflang}).

We normalize non-English efficiency by English from the same endpoint:
\[
\mathrm{Eff.\ vs.\ En}
=
\frac{\frac{1}{5}\sum_{\ell\in\{\mathrm{zh,es,fr,ar,ru}\}}
\mathrm{joint\_per\_1k}(\ell)}
{\mathrm{joint\_per\_1k}(\mathrm{en})}.
\]
Thus, $100\%$ denotes parity with English from the same checkpoint. Table~\ref{tab:progression} summarizes the complete lineage under this shared evaluation.

\begin{table*}[t]
\centering
\small
\setlength{\tabcolsep}{5.5pt}
\begin{tabular}{@{}lllccccccl@{}}
\toprule
stage & intervention & endpoint & $C@1$ & $L$ & $1{-}T$ & $J@1$ & $J@16$ & eff.\ vs En & dominant residual\\
\midrule
I & released & Qwen3-8B & $78.1\%$ & $26.9\%$ & \cellcolor{StageRed!11}$5.8\%$ & \cellcolor{StageGreen!6}$19.6\%$ & \cellcolor{StageGreen!9}$30.6\%$ & $27\%$ & language selection\\
\addlinespace[2pt]
II & multilingual SFT & + SFT (8B) & $66.0\%$ & $99.7\%$ & \cellcolor{StageRed!30}$18.2\%$ & \cellcolor{StageGreen!20}$65.1\%$ & \cellcolor{StageGreen!26}$86.3\%$ & $78\%$ & termination\\
\midrule
I & released & Qwen3-4B & $76.2\%$ & $39.8\%$ & \cellcolor{StageRed!9}$4.7\%$ & \cellcolor{StageGreen!9}$28.6\%$ & \cellcolor{StageGreen!11}$35.7\%$ & $47\%$ & language selection\\
\addlinespace[2pt]
II & multilingual SFT & + SFT & $61.4\%$ & $99.4\%$ & \cellcolor{StageRed!34}$21.1\%$ & \cellcolor{StageGreen!18}$60.2\%$ & \cellcolor{StageGreen!25}$84.6\%$ & $73\%$ & termination\\
\addlinespace[2pt]
III & correctness-only RL & RL: correctness only & $60.5\%$ & $8.7\%$ & \cellcolor{StageRed!2}$0.3\%$ & \cellcolor{StageGreen!2}$5.0\%$ & \cellcolor{StageGreen!5}$15.4\%$ & $9\%$ & returns to English\\
\addlinespace[2pt]
\rowcolor{StageGreen!10}III & target-language RL & RL: additive $\rightarrow$ gated & $60.1\%$ & $99.8\%$ & \cellcolor{StageRed!3}$0.5\%$ & \cellcolor{StageGreen!18}$60.0\%$ & \cellcolor{StageGreen!25}$81.7\%$ & $102\%$ & answer correctness\\
\bottomrule
\end{tabular}

\caption{\textbf{One protocol exposes a staged repair.}
The upper block reports the released and multilingual-SFT Qwen3-8B endpoints; the lower block reports the full Qwen3-4B lineage. Stage III has a correctness-only control and a target-language procedure whose gated
phase continues its additive phase. $C@1$, $L$, $1-T$, and $J@1$ are sample-level rates, and $J@16$ is pass@16, averaged over the five primary non-English languages. Green indicates delivery and red non-termination. \emph{Eff.\ vs En} is relative adjusted delivery efficiency, with $100\%$ denoting parity; at the released checkpoints that mean is carried by Chinese and Russian alone, because the other three languages deliver nothing. \emph{Dominant residual} identifies the remaining bottleneck.}
\label{tab:progression}
\end{table*}

\begin{table*}[t]
\centering
\small
\setlength{\tabcolsep}{5pt}
\begin{tabular}{@{}llrrrl@{}}
\toprule
run & initialized from & examples & batch & epochs (cfg.) & evaluated at\\
\midrule
\multicolumn{6}{@{}l}{\itshape supervised}\\
multilingual 4B & Qwen3-4B & 43{,}218 & 16 & 5 & epoch 2, step 5{,}402\\
multilingual 8B & Qwen3-8B & 56{,}307 & 16 & 2 & epoch 1, step 3{,}519\\
English-only & Qwen3-4B & 9{,}800 & 4 & 3 & epoch 1, step 2{,}450\\
specialist zh & Qwen3-4B & 9{,}334 & 4 & 3 & epoch 1\\
specialist es & Qwen3-4B & 9{,}491 & 4 & 3 & epoch 1\\
specialist fr & Qwen3-4B & 9{,}400 & 4 & 3 & epoch 1\\
specialist ar & Qwen3-4B & 9{,}347 & 4 & 2 & epoch 1\\
specialist ru & Qwen3-4B & 9{,}179 & 4 & 3 & epoch 1\\
\midrule
\multicolumn{6}{@{}l}{\itshape reinforcement learning, from multilingual SFT-4B except the gated continuation}\\
correctness only & multilingual 4B & 16{,}754 & 32 & --- & step 700\\
additive & multilingual 4B & 16{,}754 & 32 & --- & step 200\\
correctness-gated & additive @200 & 16{,}754 & 64 & --- & step 1{,}000\\
\bottomrule
\end{tabular}

\caption{\textbf{Complete training ledger.} The Qwen3-8B chain changes both model scale and
corpus; the English-only and specialist controls differ in batch size and data volume; and
the gated PPO run continues the additive run rather than restarting from the supervised
checkpoint. Reward curves and held-out losses for these runs appear in
Figure~\ref{fig:training}.}
\label{tab:runs}
\end{table*}

\vspace{-6pt}
\section{The Delivery Problem at Release}
\label{sec:basefindings}

\begin{figure*}[t]
\centering
\includegraphics[width=\textwidth]{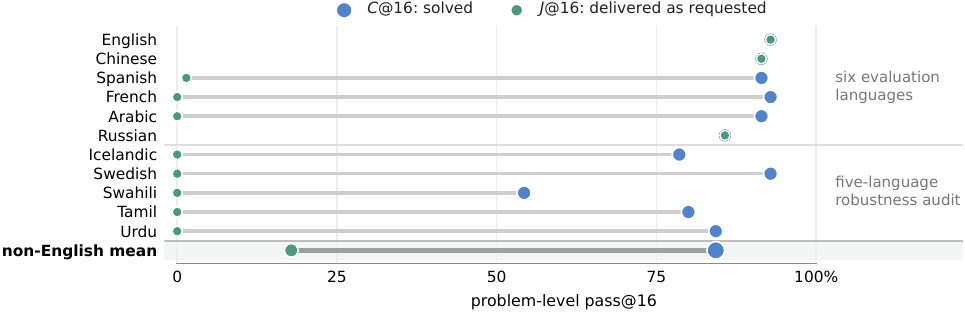}
\caption{\textbf{Released accuracy conceals failed target-language delivery.}
The figure reports $C@16$ and $J@16$ for the released Qwen3-4B on 70 problems. The emphasized row averages the ten non-English languages: five primary evaluation languages and five languages absent from all training corpora. Appendix~\ref{app:evaluation} reports problem-clustered intervals, the Qwen3-8B results, and independent-judge validation (Tables~\ref{tab:problemci} and~\ref{tab:lowres}).}
\label{fig:problem}
\end{figure*}

The released checkpoints can solve multilingual problems, but language adherence prevents most correct solutions from being delivered in the requested language. Across the five primary non-English languages, Qwen3-4B reaches $90.6\%$ $C@16$ but only $35.7\%$ $J@16$ (Figure~\ref{fig:problem}); Qwen3-8B reproduces the gap, with $92.6\%$ correctness and $30.6\%$ delivery. Both are close to English in correctness ($92.9\%$), showing that the missing capability is not primarily mathematical (per-language values and problem-clustered intervals in Table~\ref{tab:problemci}).

Language adherence, rather than termination, is the binding constraint. Mean non-English termination failure is no worse than English at either scale ($4.7\%$ versus $6.8\%$ at 4B and $5.8\%$ versus $9.2\%$ at 8B). By contrast, adherence in Spanish, French, and Arabic is at most $0.1\%$ at 4B and zero at 8B, with visible reasoning generated almost entirely in English. An independent DeepSeek-V4-Flash judge confirms this pattern, labeling $98.4\%$ of sampled Spanish, French, and Arabic traces as English and the remainder as mixed (Appendix~\ref{app:evaluation}). The behavior is not uniform across either languages or checkpoints: Chinese adheres almost perfectly at both scales, whereas Russian adheres in $99.2\%$ of samples at 4B but only $34.6\%$ at 8B. Delivery is therefore a property of the language--checkpoint pair, not of a model alone.

The delivery gap extends beyond the five primary languages. We additionally evaluate Icelandic, Swedish, Swahili, Tamil, and Urdu, which span three scripts and appear in none of our training corpora. Across all ten non-English languages, Qwen3-4B reaches $84.3\%$ $C@16$ but only $17.9\%$ $J@16$; Qwen3-8B reaches $89.0\%$ and $15.4\%$, respectively (Figure~\ref{fig:problem}). Language-specific gaps between correctness and delivery reach 94 percentage points (Swedish on Qwen3-8B: $94.3\%$ correct, none delivered).

An explicit language instruction does not reliably override the language of the problem. In a separate 30-problem probe, an English problem followed by a request to reason in Chinese, Spanish, French, or Arabic produces zero target-language adherence at the released Qwen3-4B checkpoint; Russian reaches only $4.2\%$ (Table~\ref{tab:probe}). This motivates the next intervention: training on complete target-language solutions rather than relying on an instruction alone.

\vspace{-6pt}
\section{Supervised Fine-Tuning}
\label{sec:sftfindings}

\begin{figure*}[t]
\centering
\includegraphics[width=\textwidth]{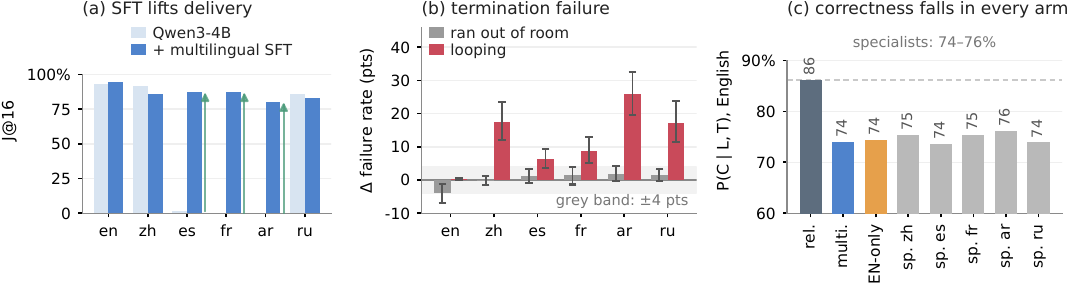}
\caption{\textbf{SFT gains delivery but exposes looping; all panels are Qwen3-4B.}
(a) $J@16$ before and after six-language SFT. (b) Paired changes in termination failure, decomposed into detected looping and token-budget exhaustion. (c) Conditional English correctness for the released checkpoint, multilingual and English-only SFT, and five single-language specialists; specialist bars denote models rather than evaluation languages. The Qwen3-8B replication of (a) and (b) is reported in Tables~\ref{tab:summary} and~\ref{tab:termination}.}
\label{fig:sft}
\end{figure*}

Multilingual SFT repairs the language-adherence bottleneck and sharply increases delivery. Across the five primary non-English languages, Qwen3-4B adherence rises from $39.8\%$ to $99.4\%$, while $J@16$ rises from $35.7\%$ to $84.6\%$ (Figure~\ref{fig:sft}a). Qwen3-8B reproduces the transition, with $J@16$ increasing from $30.6\%$ to $86.3\%$ (Table~\ref{tab:summary}). The improvement holds across all five target languages, showing that SFT converts previously English-pivoted solutions into solutions readable in the requested language.

This repair incurs an accuracy cost, but the cost is not specific to multilingual mixing. Mean non-English $C@16$ falls from $90.6\%$ to $85.1\%$ at 4B and from $92.6\%$ to $86.9\%$ at 8B. More importantly, the same decline appears in controls without multilingual mixing: conditional English correctness, pooled over samples, falls from $86.2\%$ at release to $74.1\%$ after multilingual SFT and $74.5\%$ after English-only SFT, while the five single-language specialists lie in the same $73.7\%$--$76.2\%$ range (Figure~\ref{fig:sft}c). The 8B pair reproduces the accuracy decline, although the English-only and specialist controls are available only at 4B. The evidence therefore identifies a general SFT-associated cost rather than a cost unique to multilingual training. The pattern is more consistent with displacement of prior post-training than with data quality: the released checkpoints have already undergone extensive mathematical post-training \citep{qwen3}, and further supervision on external distilled traces appears to overwrite part of it, in line with reports that limited long-chain-of-thought supervision degrades small models \citep{throughvalley2025}. Residual label noise in the source traces does not account for it, since the decline is of the same magnitude at 8B, whose corpus adds a correctness filter (Appendix~\ref{app:corpusaudit}).

SFT additionally introduces a target-language termination failure. At 4B, mean non-English termination failure rises from $4.7\%$ to $21.1\%$, while English improves from $6.8\%$ to $3.2\%$. The increase is driven by detected looping across every target language, not by exhaustion of the token budget (Figure~\ref{fig:sft}b). Qwen3-8B reproduces the same pattern: non-English termination failure rises from $5.8\%$ to $18.2\%$, while English again improves. The complete per-language decomposition and paired tests appear in Table~\ref{tab:termination}.

SFT also improves the computational efficiency of delivery. Relative non-English efficiency rises from $47\%$ to $73\%$ of English at 4B and from $27\%$ to $78\%$ at 8B. When conditioned on successful delivery, non-English traces are near parity with English; the remaining cost is carried primarily by attempts that never deliver (Appendix~\ref{app:additional}).

Thus, SFT shifts the dominant bottleneck from language adherence to termination.

\vspace{-6pt}
\section{Reinforcement Learning}
\label{sec:rlfindings}

The RL stage compares two arms trained from the same supervised checkpoint: a control that rewards correctness alone, and a target-language procedure run in two phases. Both restore termination---mean non-English termination failure falls from $21.1\%$ after SFT to $0.3\%$ and $0.5\%$---and both reach the same accuracy, $C@1=60.5\%$ and $60.1\%$.

What separates them is delivery. Without a language term, adherence collapses to $8.7\%$ and $J@16$ to $15.4\%$; under the target-language procedure adherence holds at $99.8\%$ and $J@16$ reaches $81.7\%$, while delivery efficiency rises from $73\%$ to $102\%$ of English (Table~\ref{tab:progression}). The arms are not matched in every hyperparameter---the gated phase halves the response cap and doubles the batch---but no such difference produces a 91-point gap in adherence at equal accuracy.

The procedure needs two phases because the shaping term does not stay honest. The reward ledger of its additive phase shows the language credit being collected without solving the task: among the incorrect trajectories the objective still pays, language quality rises to near $100\%$ while accuracy stays low, and they absorb $44\%$--$82\%$ of all reward paid (Figure~\ref{fig:ledger}). The gated phase withdraws that credit and continues training.

Whether the reward names the target language therefore decides whether capability is delivered in that language; gating decides whether the language credit has to be earned by solving the task.

\begin{figure*}[t]
\centering
\includegraphics[width=\textwidth]{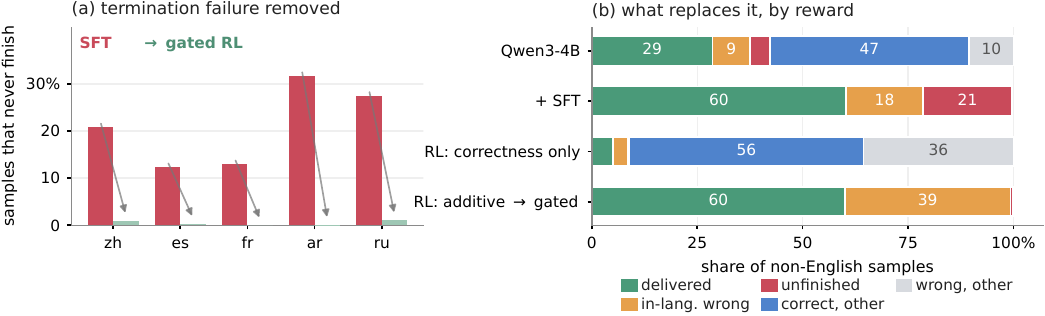}
\caption{\textbf{RL restores completion; reward design determines delivery.}
All endpoints are Qwen3-4B. The RL stage has two arms from the same multilingual SFT
checkpoint: a correctness-only control, and a target-language procedure whose gated phase
continues its additive phase; the two phase rows in (b) are checkpoints of that one run. (a) Non-termination before RL and at the end of the target-language procedure; the control and the additive phase are reported in Table~\ref{tab:progression}. (b) Response outcomes at the released and supervised checkpoints, under the control, and at the end of the target-language procedure. The reward ledger behind the phase boundary is reported in Appendix~\ref{app:additional} (Figure~\ref{fig:ledger}).}
\label{fig:rl}
\end{figure*} 

\vspace{-6pt}
\section{Analysis}
\label{sec:discussion}

The three stages form a construction rather than a sequence of unrelated failure analyses. At release, language adherence is binding: substantial correctness is hidden behind visible English reasoning. SFT repairs adherence and raises target-language delivery, after which termination becomes binding because non-English traces loop. RL repairs termination, while correctness gating prevents this repair from being exchanged for language-compliant errors. The dominant bottleneck moves because each stage removes the condition that limited the preceding one (Table~\ref{tab:progression}).

This progression also clarifies the computational cost of multilingual delivery. Adjusted non-English delivery efficiency rises from $73\%$ of English after SFT to $102\%$ at the end of the target-language procedure; when conditioned on delivered responses, raw-token efficiency reaches parity with English. The remaining end-to-end penalty is therefore carried primarily by attempts that consume tokens without delivering a solution, rather than by the length of successful non-English reasoning. Table~\ref{tab:tokeneff} gives every efficiency figure under all four accounting conventions, together with the quantity that separates them: the share of generated tokens that ends up inside a delivered solution.

\begin{table*}[t]
\centering
\small
\begin{tabular}{@{}lccccc@{}}
\toprule
& \multicolumn{2}{c}{all attempts} & \multicolumn{2}{c}{$J$-only} & \\
\cmidrule(lr){2-3}\cmidrule(lr){4-5}
stage & adjusted & raw & adjusted & raw & token yield\\
\midrule
Qwen3-4B & $47\%$ & $41\%$ & -- & -- & $34\%$\\
+ SFT & $73\%$ & $62\%$ & $110\%$ & $94\%$ & $67\%$\\
RL: correctness only & $9\%$ & $9\%$ & -- & -- & $6\%$\\
\rowcolor{StageGreen!10}RL: gated & $102\%$ & $87\%$ & $117\%$ & $100\%$ & $88\%$\\
\addlinespace[2pt]
Qwen3-8B & $27\%$ & $26\%$ & -- & -- & $23\%$\\
+ SFT (8B) & $78\%$ & $66\%$ & $111\%$ & $94\%$ & $70\%$\\
\bottomrule
\end{tabular}

\caption{\textbf{Delivery efficiency under four accounting conventions.} Efficiency is
joint successes per 1,000 generated tokens. Every entry is a percentage of the same
endpoint's English value, so $100\%$ means parity with English; the four efficiency columns
average the five primary non-English languages, and the yield column pools them. The conventions differ in which tokens are counted.
\emph{Adjusted} first divides each trace by its language's encoding factor---the same text
costs between $1.003$ times (Chinese) and $1.314$ times (Russian) as many tokens as in
English, purely because of how the tokenizer segments the script---so an adjusted column
asks whether the model produced \emph{more reasoning}, while a \emph{raw} column counts the
tokens a user is actually billed for. \emph{All attempts} charges every generated token,
including those spent on responses that never delivered; \emph{$J$-only} charges only the
tokens inside delivered solutions. The two are linked by \emph{token yield}: the share of a
language's generated tokens that ends up inside a delivered solution, here the non-English
share as a percentage of the endpoint's English share. All-attempt efficiency is $J$-only
efficiency times yield, and the identity holds in these ratios exactly as it does in the
underlying absolute shares. Absolute per-language
yields are in Appendix~\ref{app:additional} (Table~\ref{tab:yield}). $J$-only is
undefined, and omitted, where a language delivers nothing.}
\label{tab:tokeneff}
\end{table*} The Qwen3-8B results support the same interpretation (Appendix~\ref{app:additional}).

\begin{table}[t]
\centering
\scriptsize
\setlength{\tabcolsep}{2.2pt}
\begin{tabular}{@{}lccccc@{}}
\toprule
what the prompt supplies & zh & es & fr & ar & ru\\
\midrule
\multicolumn{6}{@{}l}{\itshape Qwen3-4B (released)}\\
\quad $X$ problem, asks $X$ & $99.6\%$ & $0.1\%$ & $0.0\%$ & $0.0\%$ & $99.2\%$\\
\quad English problem, asks $X$ & $0.0\%$ & $0.0\%$ & $0.0\%$ & $0.0\%$ & $4.2\%$\\
\midrule
\multicolumn{6}{@{}l}{\itshape $+$ multilingual SFT}\\
\quad $X$ problem, asks $X$ & $99.9\%$ & $99.6\%$ & $99.0\%$ & $98.4\%$ & $99.9\%$\\
\quad English problem, asks $X$ & $0.0\%$ & $2.5\%$ & $0.8\%$ & $0.0\%$ & $94.2\%$\\
\midrule
\multicolumn{6}{@{}l}{\itshape $+$ gated RL}\\
\quad $X$ problem, asks $X$ & $100.0\%$ & $99.9\%$ & $99.2\%$ & $100.0\%$ & $100.0\%$\\
\quad English problem, asks $X$ & $0.0\%$ & $1.7\%$ & $2.5\%$ & $1.7\%$ & $9.2\%$\\
\bottomrule
\end{tabular}

\caption{\textbf{The problem language is a stronger cue than the requested reasoning language.}
Entries report target-language adherence for Qwen3-4B endpoints, where X is the language of
the column (Chinese, Spanish, French, Arabic, or Russian). ``X problem, asks X'' uses the
main 70-problem evaluation; ``English problem, asks X'' uses the separate 30-problem probe.
Each cell contains 1,120 and 120 samples, respectively.}
\label{tab:probe}
\end{table}

The language-cue probe in Table~\ref{tab:probe} provides an important boundary condition. The language of the problem is a stronger cue than an instruction naming the desired reasoning language: even after SFT and gated RL, an English problem accompanied by such an instruction rarely elicits Chinese, Spanish, French, or Arabic reasoning. Russian is the exception
after SFT ($94.2\%$), where the instruction alone suffices, but gated RL returns it to
$9.2\%$. These interventions therefore obtain target-language delivery through end-to-end localized problems and solutions, rather than a general ability to switch reasoning language through instruction alone.

More broadly, multilingual post-training redistributes correlated failure modes rather than uniformly improving a single capability. Evaluating only accuracy, adherence, or termination can therefore miss both the bottleneck removed by an intervention and the failure that replaces it.

\vspace{-6pt}
\section{Limitations}\label{sec:limitations}

Our conclusions are limited to visible multilingual delivery in one Qwen3 family on competition mathematics. The released checkpoints had already undergone substantial mathematical post-training, and our interventions target delivery rather than general reasoning capability.

Each training run uses one seed, and evaluation is limited to 70 translated problems. The English-only and specialist SFT controls differ in data volume and training configuration, while the gated endpoint continues the additive run with a different response cap and batch size. These comparisons are therefore diagnostic rather than fully matched causal ablations.

Finally, although the prompts were constructed in parallel and the translations underwent extensive quality audits, translation can still alter the naturalness or difficulty of mathematical problems and reasoning instructions.

\vspace{-6pt}
\section{Conclusion}\label{sec:conclusion}
We studied when post-training makes multilingual reasoning usable rather than merely
accurate. Across thirteen endpoints of one Qwen3 lineage, a fixed response-level readout
shows that released checkpoints often solve non-English problems through visible English
reasoning, causing answer accuracy to overstate target-language delivery.
Following this lineage reveals a shifting bottleneck. Multilingual SFT repairs language
adherence, but its general accuracy cost is accompanied by a target-language termination
failure. RL restores termination under both arms and at equal accuracy; what the language
term decides---and only when it is gated on correctness---is whether the recovered responses
are delivered in the requested language or return to English. Together, these stages establish a constructive post-training path from English-pivoted
capability to reliable multilingual delivery, converting capability already present at
release into correct and complete target-language solutions with efficiency comparable to
English.

\begin{small}
\bibliography{references}
\end{small}

\onecolumn
\section*{Supplementary Material}
\addcontentsline{toc}{section}{Supplementary Material}

\appendix
\setcounter{secnumdepth}{2}
\renewcommand{\thetable}{S\arabic{table}}
\renewcommand{\thefigure}{S\arabic{figure}}
\setcounter{table}{0}
\setcounter{figure}{0}
\section{Evaluation Protocol and Instrument Checks}\label{app:evaluation}

All primary experiments use 70 competition-mathematics problems (30 AIME-2026 and 40
AMC-2023), six fully localized prompts, 16 samples per problem at temperature $0.7$, and a
24,576-token response budget; the five single-language specialists are evaluated in two
languages only, their own and English. The five-language mean in the main SFT and RL results means
Chinese, Spanish, French, Arabic, and Russian; English is a control language. The released-
checkpoint robustness audit additionally evaluates Icelandic, Swedish, Swahili, Tamil, and
Urdu, but no fine-tuned endpoint is trained or evaluated on those five languages.

\paragraph{Correctness.}
We extract only the final brace-matched \verb|\boxed{}| answer or the text after the last
English, Chinese, Spanish, French, Arabic, or Russian final-answer marker. The extracted
answer, never the full trace, is checked with \texttt{math\_verify}; a missing final answer
is incorrect. This prevents a truncated trace that merely mentions the gold value from
receiving credit.

\paragraph{Language adherence.}
We extract the visible \texttt{<think>} block when present, remove mathematical expressions,
split the remaining text into sentence-like segments, and evenly subsample at most 60
segments so that both the beginning and end of a long trace are represented. \texttt{langid}
is restricted to the six evaluation languages. $L=1$ when at least half of the segments are
classified as the requested language; English uses the identical definition with English
as its target. Script fraction is retained as a model-free cross-check. A blind LLM judge
re-scores the same 609 non-English traces and agrees with the continuous target-language
fraction that $L$ thresholds at Pearson $r=97.9\%$ (mean absolute difference $5.7\%$).

\paragraph{Termination.}
$T=1$ requires both that generation does not reach the response cap and that its
language-agnostic repetition score is at most $0.85$. The score is
$1-\text{compressed bytes}/\text{raw bytes}$ using zlib compression. We report cap-only and
looping components separately. A blind judge reclassifies 170 stratified failures; the
cap-overrun versus looping distinction reaches $89.4\%$ accuracy and Cohen's $\kappa=.519$.
Finer loop taxonomies are not used quantitatively.

$C@k$ and $J@k$ use the standard unbiased pass@$k$ estimator with $C$ or $J$ as
the success event. Intervals resample problems as blocks ($B=2{,}000$), because samples of a
problem are not independent (Table~\ref{tab:problemci}); paired comparisons use the same problem--language cells and
apply Holm correction within each five-language family.

\subsection{Released-checkpoint results and uncertainty}

\begin{table}[!htb]
\centering
\small
\setlength{\tabcolsep}{2.5pt}
\begin{tabular}{llccccccc}
\toprule
checkpoint & language & $C$@16 [95\% CI] & $J$@16 [95\% CI] & $L$ & $1{-}T$ & cap only & looping & tokens\\
\midrule
Qwen3-4B & English & $92.9\%$ [85.7, 98.6] & $92.9\%$ [85.7, 98.6] & $100.0\%$ & $6.8\%$ & $6.7\%$ & $0.1\%$ & 9,568\\
 & Chinese & $91.4\%$ [84.3, 97.1] & $91.4\%$ [84.3, 97.1] & $99.6\%$ & $3.4\%$ & $2.5\%$ & $0.9\%$ & 8,127\\
 & Spanish & $91.4\%$ [84.3, 97.1] & $1.4\%$ [0.0, 4.3] & $0.1\%$ & $4.7\%$ & $4.4\%$ & $0.4\%$ & 7,928\\
 & French & $92.9\%$ [85.7, 98.6] & $0.0\%$ [0.0, 0.0] & $0.0\%$ & $2.7\%$ & $2.7\%$ & $0.0\%$ & 7,158\\
 & Arabic & $91.4\%$ [84.3, 97.1] & $0.0\%$ [0.0, 0.0] & $0.0\%$ & $3.8\%$ & $3.5\%$ & $0.4\%$ & 7,613\\
 & Russian & $85.7\%$ [77.1, 92.9] & $85.7\%$ [77.1, 92.9] & $99.2\%$ & $8.8\%$ & $3.1\%$ & $5.6\%$ & 8,380\\
 & Icelandic & $78.6\%$ [68.6, 87.1] & $0.0\%$ [0.0, 0.0] & $0.0\%$ & $5.9\%$ & $4.8\%$ & $1.1\%$ & 8,691\\
 & Swedish & $92.9\%$ [85.7, 98.6] & $0.0\%$ [0.0, 0.0] & $0.0\%$ & $4.2\%$ & $4.0\%$ & $0.2\%$ & 7,759\\
 & Swahili & $54.3\%$ [42.9, 67.1] & $0.0\%$ [0.0, 0.0] & $0.0\%$ & $6.2\%$ & $3.5\%$ & $2.8\%$ & 8,983\\
 & Tamil & $80.0\%$ [70.0, 88.6] & $0.0\%$ [0.0, 0.0] & $0.0\%$ & $3.7\%$ & $2.3\%$ & $1.3\%$ & 7,244\\
 & Urdu & $84.3\%$ [75.7, 92.9] & $0.0\%$ [0.0, 0.0] & $0.0\%$ & $2.6\%$ & $1.7\%$ & $0.9\%$ & 7,053\\
\midrule
Qwen3-8B & English & $92.9\%$ [85.7, 98.6] & $92.9\%$ [85.7, 98.6] & $99.9\%$ & $9.2\%$ & $9.0\%$ & $0.2\%$ & 10,446\\
 & Chinese & $91.4\%$ [84.3, 97.1] & $91.4\%$ [84.3, 97.1] & $99.9\%$ & $7.5\%$ & $4.5\%$ & $3.0\%$ & 9,790\\
 & Spanish & $94.3\%$ [88.6, 98.6] & $0.0\%$ [0.0, 0.0] & $0.0\%$ & $4.3\%$ & $4.2\%$ & $0.1\%$ & 8,493\\
 & French & $92.9\%$ [87.1, 98.6] & $0.0\%$ [0.0, 0.0] & $0.0\%$ & $4.9\%$ & $4.9\%$ & $0.0\%$ & 8,756\\
 & Arabic & $92.9\%$ [85.7, 98.6] & $0.0\%$ [0.0, 0.0] & $0.0\%$ & $5.1\%$ & $5.1\%$ & $0.0\%$ & 8,494\\
 & Russian & $91.4\%$ [84.3, 97.1] & $61.4\%$ [50.0, 72.9] & $34.6\%$ & $7.4\%$ & $5.7\%$ & $1.7\%$ & 9,393\\
 & Icelandic & $87.1\%$ [78.6, 94.3] & $0.0\%$ [0.0, 0.0] & $0.0\%$ & $7.9\%$ & $6.9\%$ & $1.0\%$ & 9,744\\
 & Swedish & $94.3\%$ [88.6, 98.6] & $0.0\%$ [0.0, 0.0] & $0.0\%$ & $5.6\%$ & $5.5\%$ & $0.1\%$ & 9,041\\
 & Swahili & $70.0\%$ [58.6, 80.0] & $1.4\%$ [0.0, 4.3] & $1.0\%$ & $11.3\%$ & $9.1\%$ & $2.1\%$ & 10,726\\
 & Tamil & $85.7\%$ [77.1, 92.9] & $0.0\%$ [0.0, 0.0] & $0.0\%$ & $4.8\%$ & $4.2\%$ & $0.6\%$ & 8,858\\
 & Urdu & $90.0\%$ [82.9, 95.7] & $0.0\%$ [0.0, 0.0] & $0.0\%$ & $4.8\%$ & $4.0\%$ & $0.8\%$ & 8,479\\
\bottomrule
\end{tabular}

\caption{Accuracy and joint delivery with 95\% problem-clustered bootstrap intervals. The
two termination columns are budget exhaustion without repetition and detected looping; a
trace can satisfy both conditions. ``tokens'' is the mean over all attempts.}
\label{tab:problemci}
\end{table}

\subsection{Robustness beyond the primary languages}

The additional languages span Latin, Tamil, and Arabic scripts and appear in none of the
corpora we construct (Table~\ref{tab:lowres}). The language identifier and blind judge agree that visible
target-language reasoning is essentially absent in all ten released-checkpoint cells; the
only nonzero $J@16$ is Swahili on Qwen3-8B ($1.4\%$).

\begin{table}[!htb]
\centering
\small
\setlength{\tabcolsep}{4pt}
\begin{tabular}{@{}llccc@{\hspace{1.1em}}cccc@{}}
\toprule
 & & \multicolumn{3}{c}{\emph{outcome}} & \multicolumn{4}{c}{\emph{two instruments for $L$}} \\
\cmidrule(lr){3-5}\cmidrule(l){6-9}
model & lang & $C@16$ & $J@16$ & gap & langid-$L$ & judge-en & judge-tgt & judge-frac \\
\midrule
\texttt{Qwen3-4B} & is & $78.6\%$ & $0.0\%$ & $78.6\%$ & $0.0\%$ & $84.4\%$ & $0.0\%$ & $2.0\%$ \\
 & sv & $92.9\%$ & $0.0\%$ & $92.9\%$ & $0.0\%$ & $96.9\%$ & $0.0\%$ & $0.5\%$ \\
 & sw & $54.3\%$ & $0.0\%$ & $54.3\%$ & $0.0\%$ & $78.1\%$ & $0.0\%$ & $1.9\%$ \\
 & ta & $80.0\%$ & $0.0\%$ & $80.0\%$ & $0.0\%$ & $93.8\%$ & $0.0\%$ & $0.8\%$ \\
 & ur & $84.3\%$ & $0.0\%$ & $84.3\%$ & $0.0\%$ & $90.6\%$ & $0.0\%$ & $1.0\%$ \\
\midrule
\texttt{Qwen3-8B} & is & $87.1\%$ & $0.0\%$ & $87.1\%$ & $0.0\%$ & $93.8\%$ & $0.0\%$ & $1.2\%$ \\
 & sv & $94.3\%$ & $0.0\%$ & $94.3\%$ & $0.0\%$ & $100.0\%$ & $0.0\%$ & $0.2\%$ \\
 & sw & $70.0\%$ & $1.4\%$ & $68.6\%$ & $1.0\%$ & $90.6\%$ & $0.0\%$ & $0.3\%$ \\
 & ta & $85.7\%$ & $0.0\%$ & $85.7\%$ & $0.0\%$ & $96.9\%$ & $0.0\%$ & $0.3\%$ \\
 & ur & $90.0\%$ & $0.0\%$ & $90.0\%$ & $0.0\%$ & $93.8\%$ & $0.0\%$ & $0.5\%$ \\
\bottomrule
\end{tabular}

\caption{Robustness audit over five additional languages, none of which appears in any
corpus we construct; \emph{gap} is $C@16-J@16$. The right-hand block reports two
independently constructed instruments for $L$. \emph{langid-$L$} is the automatic
per-response criterion, with the langid candidate set extended to all eleven evaluation
languages so that these five can be detected at all. \emph{judge-en} and \emph{judge-tgt}
are the shares of sampled traces a blind LLM judge labels as reasoning in English and in the
target language, and \emph{judge-frac} is its estimate of how much of the trace is written
in the target language. The two instruments agree in all ten cells, and \emph{judge-en} is a
positive identification of English rather than a failure to identify the target language, so
the absent target-language reasoning is a property of the models and not an artifact of
either instrument.}
\label{tab:lowres}
\end{table}

\subsection{Input-language versus instruction probe}

The probe table in the main text reports the outcome; here we record its protocol and the
measurements behind it. This probe is a separate experiment, not a slice of the main
evaluation. Each of the seven runs draws its own 30 problems (15 AIME-2026 and 15 AMC-23)
and covers six languages at $K=4$, giving 720 samples per run. Only the runner, token budget
and scoring are shared, so the probe's absolute values are not comparable with the endpoint
tables and are never pooled with them; what transfers is the within-probe contrast between
conditions. The forward set states the problem in English and appends a localized
instruction to reason in the target language.

Table~\ref{tab:e6} gives the underlying quantities for the supervised checkpoint under the
forward condition. Correctness is essentially flat across probe conditions at every
endpoint ($47\%$--$72\%$ after SFT, $71\%$--$86\%$ at release), so refusal to switch language
is not a failure to solve. Runs on the released Qwen3-8B, and a reverse set that states the
problem in the target language and asks for English, are released with the repository but
not reported here; the table covers the Qwen3-4B lineage under the forward condition only.

\begin{table}[!htb]
\centering
\small
\begin{tabular}{@{}lcccc@{}}
\toprule
instructed language & $n$ & $L$ & target frac. & $C$ \\
\midrule
en & 120 & $100.0\%$ & $91.3\%$ & $56.7\%$ \\
zh & 120 & $0.0\%$ & $5.7\%$ & $59.2\%$ \\
es & 120 & $2.5\%$ & $3.6\%$ & $59.2\%$ \\
fr & 120 & $0.8\%$ & $1.7\%$ & $56.7\%$ \\
ar & 120 & $0.0\%$ & $0.3\%$ & $54.2\%$ \\
ru & 120 & $94.2\%$ & $86.7\%$ & $46.7\%$ \\
\bottomrule
\end{tabular}

\caption{English problem text with an instruction to reason in the listed language,
multilingual SFT on Qwen3-4B. $L$ measures the language of the visible solution trace; ``target
frac.'' is its mean instructed-language fraction.}
\label{tab:e6}
\end{table}

\section{Data, Training, and Reproducibility}\label{app:training}

\subsection{Supervised corpus construction}\label{app:corpus}

The SFT corpus begins with 45K English OpenR1 long-chain-of-thought mathematics traces.
Traces are assigned languages round robin by global index; English traces are retained and
the other five are translated by a locally served Qwen3-14B. Figures and display
environments are detached before translation and reattached afterward; \texttt{<think>}
blocks are preserved; and long traces are chunked. An alignment gate rejects any trace with
a capped chunk or a changed number or value of \verb|\boxed| answers. Seven further gates
then run on each translated trace: exactly one \texttt{<think>} and one \texttt{</think>};
a recoverable \verb|\boxed| answer; an even number of dollar signs; matched braces; a
bounded ratio of translated to source length; and no repetition, flagged when zlib
compresses the trace by more than $70\%$. The seventh gate is target-language fidelity, and
it branches by script, because the same test cannot serve both kinds of language. Chinese,
Arabic and Russian are written in scripts English does not use, so the gate measures the
share of prose characters in the target script and requires it to exceed $0.5$. Spanish and
French share the Latin alphabet with English, where that measure carries no information, so
the gate instead measures the share of English function words and requires it to stay below
$0.15$. The same asymmetry recurs when we score adherence at evaluation time.
The resulting 4B multilingual corpus has 43,218 traces (6,976--7,214 per translated language, 7,460 English); the 8B
replication uses a separately constructed 56,307-trace corpus.

\subsection{Translator selection}\label{app:translator}

A comparison whose criteria and pass marks were fixed before any output was scored favours
Qwen3-14B on every axis. Qwen3-14B and Hunyuan-MT-7B translated the same
40 source traces into all five languages under the same chunking, and were judged on
three independent axes.

The primary axis is mechanical and uses no judge. Over 2,515 aligned chunks per arm,
Qwen3-14B preserves every \verb|\boxed| answer and meets the pre-registered $.98$ number-recall
mark in all five languages; Hunyuan recall is $.875$--$.939$ and fails it everywhere
(Table~\ref{tab:translator}). Losing digits is disqualifying for mathematical data
regardless of how fluent the prose is. The second axis is reproducibility: on two
independently drawn Chinese samples Qwen3-14B recalls $.997$ and $.998$ of numbers while
Hunyuan recalls $.939$ and $.935$, so the gap is a stable property rather than sampling
noise. The third axis is a blind 250-pair judge comparison, which scores faithfulness at
$4.85$ against $2.91$ on a five-point scale with a 209:19:22 win/loss/tie record.

The three axes agree, which matters more than any one of them. The one blemish is that
Qwen3-14B leaves English discourse markers in Chinese output, giving a target-script purity
of $.877$; the judge did not penalize this as unfaithful (Chinese faithfulness $4.90$) and
number recall is unaffected. This comparison selects a data-preparation tool, not a model
under study.

\begin{table}[!htb]
\centering
\small
\setlength{\tabcolsep}{4.5pt}
\begin{tabular}{@{}llccccc@{}}
\toprule
 &  & \multicolumn{4}{c}{mechanical fidelity} & blind\\
\cmidrule(lr){3-6}
language & translator & boxed & digits & \LaTeX & trunc. & faithfulness\\
\midrule
Chinese & Qwen3-14B & $100.0\%$ & $99.7\%$ & $100.0\%$ & $0.0\%$ & $4.90$\\
 & Hunyuan-MT-7B & $97.0\%$ & $93.9\%$ & $100.0\%$ & $0.4\%$ & $3.04$\\
\addlinespace[1pt]
Spanish & Qwen3-14B & $100.0\%$ & $100.0\%$ & $100.0\%$ & $0.0\%$ & $4.88$\\
 & Hunyuan-MT-7B & $96.6\%$ & $91.8\%$ & $100.0\%$ & $0.0\%$ & $3.12$\\
\addlinespace[1pt]
French & Qwen3-14B & $100.0\%$ & $99.9\%$ & $100.0\%$ & $0.0\%$ & $4.96$\\
 & Hunyuan-MT-7B & $95.4\%$ & $91.9\%$ & $100.0\%$ & $0.0\%$ & $2.96$\\
\addlinespace[1pt]
Arabic & Qwen3-14B & $100.0\%$ & $98.9\%$ & $100.0\%$ & $0.2\%$ & $4.54$\\
 & Hunyuan-MT-7B & $95.4\%$ & $87.5\%$ & $100.0\%$ & $0.0\%$ & $2.84$\\
\addlinespace[1pt]
Russian & Qwen3-14B & $100.0\%$ & $99.6\%$ & $100.0\%$ & $0.0\%$ & $4.98$\\
 & Hunyuan-MT-7B & $95.2\%$ & $89.6\%$ & $100.0\%$ & $0.0\%$ & $2.58$\\
\addlinespace[1pt]
\midrule
all & Qwen3-14B & \multicolumn{4}{c}{all criteria met} & $\mathbf{4.85}$\\
all & Hunyuan-MT-7B & \multicolumn{4}{c}{digit recall below threshold} & $2.91$\\
\bottomrule
\end{tabular}

\caption{Pre-registered translator comparison, per language. Mechanical columns are over
2,515 aligned chunks per arm; faithfulness is a blind judge score over 250 randomized
pairs. The pre-specified number-recall pass mark was $.98$.}
\label{tab:translator}
\end{table}

\subsection{Fidelity of the corpus we trained on}\label{app:corpusaudit}

The bake-off certifies the tool; the corpus that reaches the optimizer is audited
separately, because a study whose central finding is an English pivot must show that the
pivot is not an artefact of English left in its own training data. We sample 250 traces per
language from the 4B corpus and score them mechanically and with an independent judge
(Table~\ref{tab:corpus}).

Structure survives translation: $99.6$--$100\%$ of sampled traces still reach a
\verb|\boxed| answer and $98.8$--$100\%$ keep their mathematics intact. Reasoning quality is unchanged in the sense
that matters here: judged coherence is $4.47$--$4.90$ per language against $4.88$ for the
untranslated English source, so translation did not degrade the arguments it carried.
Target-language fidelity is $4.32$--$4.62$ except in Arabic, at $3.54$.

The residual is English leakage, and it is the confound worth naming. Judged moderate or
severe leakage affects $4.0$--$6.0\%$ of Chinese, Spanish, French, and Russian traces and
$16.0\%$ of Arabic ones. The corpus built for the 8B replication was re-translated and is
uniformly cleaner, most where the first was worst (Arabic fidelity $3.54\rightarrow4.12$,
moderate-or-severe leakage $.160\rightarrow.120$). Two observations bound the risk. The
leakage is an upper bound on what a model could imitate, and the model trained on this
corpus does not imitate it: post-SFT adherence is $.984$--$.999$, so a corpus that is a few
percent contaminated did not produce a few percent of English traces. And leakage cannot
explain the released-checkpoint result at all, which is measured before any of this data
exists.

Two further properties of the source are inherited rather than introduced by us. The
OpenR1 traces were filtered structurally but not for answer correctness, and a
judged sample puts the wrong-final-answer rate at roughly $18.5\%$. This rate is identical
in the English-only control and in the multilingual corpus, since both are built from the
same pool, so it is common to both arms of every language comparison we draw and cannot
produce a difference between them; it does place a ceiling on the absolute accuracy any
model fine-tuned on this data can reach. The 8B corpus adds a correctness filter and
8-gram decontamination against the evaluation sets.

\begin{table}[!htb]
\centering
\small
\setlength{\tabcolsep}{4pt}
\begin{tabular}{@{}lccccc@{}}
\toprule
language & coherence & fidelity & reaches \verb|\boxed| & math kept & no major leak\\
\midrule
Chinese & $4.90$ & $4.62$ & $99.6\%$ & $100.0\%$ & $96.0\%$\\
Spanish & $4.78$ & $4.40$ & $99.6\%$ & $100.0\%$ & $94.0\%$\\
French & $4.81$ & $4.32$ & $99.6\%$ & $100.0\%$ & $95.6\%$\\
Arabic & $4.47$ & $3.54$ & $99.6\%$ & $99.6\%$ & $84.0\%$\\
Russian & $4.76$ & $4.40$ & $100.0\%$ & $98.8\%$ & $94.4\%$\\
\midrule
English (untranslated) & $4.88$ & --- & $100.0\%$ & $100.0\%$ & ---\\
\bottomrule
\end{tabular}

\caption{Audit of the 4B multilingual corpus, 250 sampled traces per language. Coherence
and target-language fidelity are independent judge scores on a five-point scale; the last
column is the share of traces free of moderate or severe English residue. The English row
is the untranslated source, i.e. the floor these scores are read against.}
\label{tab:corpus}
\end{table}

\subsection{Training recipes}

Both stages run on verl; the training curves and the run ledger appear in the main paper. Every configured value in Tables~\ref{tab:recipesft} and~\ref{tab:reciperl}
is read from the configuration echoed at the start of each run's log, so the tables
describe the runs that produced the checkpoints rather than the launch scripts.

\paragraph{Supervised fine-tuning.}
We fine-tune all parameters of the released checkpoints on the corpora of
Appendix~\ref{app:corpus}, using the Qwen3 chat template with thinking enabled and
minimizing the token-level negative log-likelihood over response tokens;
Table~\ref{tab:recipesft} lists every configured value. Two divergences matter for reading
the results. The two multilingual runs use batch 16 on 16 GPUs while the English-only and
specialist controls use batch 4 on 4 GPUs. And the cutoff falls on Arabic, the language
with the worst post-SFT termination: every run uses 32,768 tokens except the Arabic
specialist, which uses 24,576. Configured epochs and
evaluated checkpoints differ by run and are listed in the run ledger in the main paper.

\paragraph{Reinforcement learning.}
PPO with GAE advantages, initialized from the multilingual SFT checkpoint on Qwen3-4B and
trained on the localized DAPO-Math-17K prompts. There is no KL term in either the reward or
the loss, so the two reward definitions given in the main text are the only shaping the
policy receives beyond the overlong-response penalty, which is identical across variants.
Because $\gamma=\lambda=1$ and the reward is a single terminal scalar, the advantage reduces
to $R(\tau)-V_\phi(s_t)$, which is the quantity the reward-geometry analysis reads. The
correctness-only and additive runs share a configuration; the gated run continues the
additive run from step 200 with a halved response cap and doubled batch, so it is a
continuation rather than a matched arm.
Training-time validation is not comparable with the endpoint evaluation and is not reported
as a result. It scores AIME-2026 alone, without the 40 AMC-23 problems, under the training
response cap rather than the 24,576-token evaluation budget, and at two or four rollouts per
prompt rather than $K=16$. All three depress it relative to the endpoint numbers; it is used
only to confirm that a run is learning, not to measure it.

\begin{table}[!htb]
\centering
\small
\setlength{\tabcolsep}{4pt}
\begin{tabular}{@{}ll@{}}
\toprule
Hyperparameter & Value\\
\midrule
Framework & verl SFT trainer, FSDP\\
Base models & Qwen3-4B, Qwen3-8B (released, post-trained)\\
Chat template & Qwen3, thinking enabled\\
Sequence cutoff length & 32{,}768 (Arabic specialist: 24{,}576)\\
Padding & no\_padding\\
Truncation & left\\
Train batch size & 16 (multilingual) / 4 (controls)\\
Micro batch per GPU & 1\\
Dynamic batching & True\\
Tokens per GPU & 98304 / 73728\\
Optimizer & AdamW, $\beta=(0.9,0.999)$\\
Learning rate & $2\times10^{-5}$, cosine to $0.1\times$ peak\\
Warmup ratio & 0.01\\
Weight decay & 0.1\\
Gradient clip & 1.0\\
Precision & bfloat16\\
Epochs configured & 5 (4B) / 2 (8B) / 3 (English-only) / 2--3 (specialists)\\
Hardware & A100-SXM4-40GB; 16 GPUs, 4 nodes (multilingual) / 4 GPUs, 1 node (controls)\\
\bottomrule
\end{tabular}

\caption{Key training hyperparameters used in supervised fine-tuning.}
\label{tab:recipesft}
\end{table}

\begin{table}[!htb]
\centering
\small
\setlength{\tabcolsep}{4pt}
\begin{tabular}{@{}ll@{}}
\toprule
Hyperparameter & Value\\
\midrule
Framework & verl, FSDP\\
Algorithm & PPO, GAE advantages\\
Initialized from & multilingual SFT on Qwen3-4B / additive @ step 200 (gated)\\
Training dataset & DAPO-Math-17K, localized (16{,}754 prompts; 16{,}751 after the over-length filter)\\
Max prompt length & 2048\\
Max response length & 16384 / 8192 (gated)\\
Overlong buffer & 2{,}048 (soft penalty, factor 1.0)\\
KL in reward / loss & none / none\\
Clip ratio & 0.2\\
Discount $\gamma$, GAE $\lambda$ & 1.0, 1.0\\
Advantage normalization & masked whitening over the batch (GAE path)\\
Entropy coefficient & 0\\
Learning rate (actor) & $1\times10^{-6}$\\
Learning rate (critic) & $1\times10^{-5}$\\
LR warmup steps & 10\\
Gradient clip & 1.0\\
Loss aggregation & token mean\\
Sampling temperature & 1.0 (train), 0.7 (validation)\\
Top-$p$ / top-$k$ & 1.0 / $-1$\\
Rollouts per prompt & 8 (train); 2 ($C$ only) / 4 (additive, gated) (validation)\\
Train batch size & 32 / 64 (gated)\\
PPO mini-batch size & 32 / 64 (gated)\\
Max steps & 1{,}500 (evaluated steps in the main paper's run ledger)\\
Hardware & 8 GPUs, 1 node, tensor-parallel 2 for rollout\\
\bottomrule
\end{tabular}

\caption{Key training hyperparameters used in reinforcement learning. Values separated by a
slash differ between the correctness-only and additive runs on one side and the gated
continuation on the other.}
\label{tab:reciperl}
\end{table}

\begin{figure}[!htb]
\centering
\includegraphics[width=0.62\textwidth]{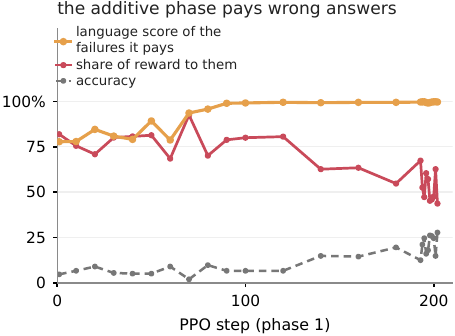}
\caption{The reward ledger of the additive phase, recorded on training rollouts of the
Qwen3-4B target-language procedure. Among the incorrect trajectories the objective still
pays, the implied language score rises to near $100\%$ while their accuracy stays low, and
they absorb $44\%$--$82\%$ of all reward paid across the recorded steps, peaking at $92\%$
early in the phase. The gated phase removes this by construction,
which is why the procedure switches. Markers are the recorded steps; the last ten are
consecutive.}
\label{fig:ledger}
\end{figure}

\subsection{Reproducibility statement}

All generated traces and per-sample measurements are retained in the accompanying artifact.
The analysis scripts are independent, parameter-free entry points with machine-readable
outputs and run logs.
The staged progression table, the lineage delivery table and the summary figure come from
A20; the released-checkpoint figure and its interval table from A15; the termination
decomposition from A14; the two analysis-section figures from A17; the training-diagnostics
figure from A16; the probe table from A18; the translation and difficulty tables from A25;
the decontamination table from A26; the recipe tables from A27; the efficiency conventions,
per-language efficiency and token-yield tables from A29, which also recomputes A11's
encoding-adjusted content cost as a regression check; the efficiency figure from
\texttt{build\_figures.py}; the remaining shared tables from
\texttt{build\_paper\_tables.py}; and the case page from A23. The one exception is the
training ledger in the main paper, which is transcribed by hand from the run metadata.
LLM-judge outputs are cached, and no item is removed from the fixed
70-problem evaluation set. The case page is rendered outside \LaTeX{} because pdflatex has
no Chinese or Arabic font on our system; A23 embeds Noto Sans CJK SC and Noto Naskh Arabic
(SIL Open Font License) and reshapes Arabic before drawing it.

\section{Additional Results}\label{app:additional}

\subsection{SFT controls and detailed paired results}

Table~\ref{tab:sftcontrols} summarizes the controls, Table~\ref{tab:sftpaired} gives the
per-language paired deltas behind them, and Table~\ref{tab:termination} separates the
termination change into looping and budget exhaustion.

\begin{table}[!htb]
\centering
\small
\begin{tabular}{@{}lrrr@{}}
\toprule
comparison & $\Delta J@1\,(\mathrm{NE})$ & $\Delta C_{\mathrm{en}}$ & $\Delta{(1-T)}_{\mathrm{NE}}$ \\
\midrule
Qwen3-4B\,$\to$\,SFT & $+31.6\%$ & $-8.8\%$ & $+16.4\%$ \\
Qwen3-8B\,$\to$\,SFT & $+45.5\%$ & $-6.8\%$ & $+12.4\%$ \\
Qwen3-4B\,$\to$\,English-only & $-5.7\%$ & $-9.2\%$ & $+4.0\%$ \\
\midrule
mixed\,$-$\,spec. (own) & $[-0.03,+0.03]$ & -- & -- \\
mixed\,$-$\,spec. (en) & -- & $[+0.01,+0.05]$ & -- \\
\bottomrule
\end{tabular}

\caption{Compact SFT control summary. NE denotes the five primary non-English languages.
The mixed--specialist rows compare the Qwen3-4B multilingual SFT endpoint with the five
Qwen3-4B single-language specialists, paired on each specialist's own language or on
English.}
\label{tab:sftcontrols}
\end{table}

\begin{table}[!htb]
\centering
\small
\begin{tabular}{@{}llrrr@{}}
\toprule
edge & lang & $\Delta J@1$ & $\Delta C$ & $\Delta$term.\ fail. \\
\midrule
Qwen3-4B $\to$ multilingual SFT & en & $-8.7\%$ & $-8.8\%$ & $-3.6\%$ \\
 & zh & $-16.5\%^{*}$ & $-16.6\%^{*}$ & $+17.4\%^{*}$ \\
 & es & $+64.5\%^{*}$ & $-14.8\%^{*}$ & $+7.6\%^{*}$ \\
 & fr & $+64.5\%^{*}$ & $-13.9\%^{*}$ & $+10.3\%^{*}$ \\
 & ar & $+53.6\%^{*}$ & $-22.1\%^{*}$ & $+27.9\%^{*}$ \\
 & ru & $-7.9\%^{*}$ & $-6.2\%^{*}$ & $+18.8\%^{*}$ \\
\midrule
Qwen3-8B $\to$ multilingual SFT (8B) & en & $-6.2\%$ & $-6.8\%$ & $-6.3\%$ \\
 & zh & $-10.7\%^{*}$ & $-10.8\%^{*}$ & $+11.1\%^{*}$ \\
 & es & $+68.4\%^{*}$ & $-11.3\%^{*}$ & $+4.9\%^{*}$ \\
 & fr & $+67.1\%^{*}$ & $-11.9\%^{*}$ & $+8.6\%^{*}$ \\
 & ar & $+60.2\%^{*}$ & $-15.4\%^{*}$ & $+24.2\%^{*}$ \\
 & ru & $+42.2\%^{*}$ & $-11.2\%^{*}$ & $+13.0\%^{*}$ \\
\midrule
Qwen3-4B $\to$ English-only SFT & en & $-9.0\%$ & $-9.2\%$ & $-2.6\%$ \\
 & zh & $-13.1\%^{*}$ & $-13.1\%^{*}$ & $+8.2\%^{*}$ \\
 & es & $-0.1\%$ & $-13.0\%^{*}$ & $-1.2\%$ \\
 & fr & $+0.0\%$ & $-11.1\%^{*}$ & $+0.1\%$ \\
 & ar & $+0.0\%$ & $-10.7\%^{*}$ & $-1.7\%$ \\
 & ru & $-15.1\%^{*}$ & $-14.9\%^{*}$ & $+14.5\%^{*}$ \\
\bottomrule
\end{tabular}

\caption{Per-language paired deltas for the three SFT-side comparisons
($^{*}$: significant after Holm correction). The English-default languages drive the
delivery gains, while correctness declines in English as well as the target languages.}
\label{tab:sftpaired}
\end{table}

\begin{table}[!htb]
\centering
\small
\begin{tabular}{lcccccc}
\toprule
 & English & Chinese & Spanish & French & Arabic & Russian\\
\midrule
\multicolumn{7}{@{}l}{\itshape Qwen3-4B $\rightarrow$ multilingual SFT-4B}\\
\quad looping & $+0.002$ & $+0.175$$^{*}$ & $+0.062$$^{*}$ & $+0.088$$^{*}$ & $+0.259$$^{*}$ & $+0.173$$^{*}$\\
\quad ran out of room & $-0.037$ & $-0.001$ & $+0.013$ & $+0.014$ & $+0.020$ & $+0.014$\\
\multicolumn{7}{@{}l}{\itshape Qwen3-8B $\rightarrow$ multilingual SFT-8B}\\
\quad looping & $+0.004$ & $+0.121$$^{*}$ & $+0.047$$^{*}$ & $+0.082$$^{*}$ & $+0.258$$^{*}$ & $+0.141$$^{*}$\\
\quad ran out of room & $-0.067$ & $-0.011$ & $+0.002$ & $+0.004$ & $-0.016$ & $-0.011$\\
\multicolumn{7}{@{}l}{\itshape Qwen3-4B $\rightarrow$ English-only SFT-4B}\\
\quad looping & $+0.003$ & $+0.080$$^{*}$ & $+0.001$ & $+0.008$$^{*}$ & $+0.000$ & $+0.149$$^{*}$\\
\quad ran out of room & $-0.029$ & $+0.002$ & $-0.013$ & $-0.007$ & $-0.017$ & $-0.004$\\
\bottomrule
\end{tabular}

\caption{Paired changes in termination failures. Asterisks mark Holm-corrected significance
within each five-language comparison family.}
\label{tab:termination}
\end{table}

\subsection{Is a fluent wrong answer better than a loop?}
The outcome composition in the main text invites the objection: delivered samples stay at $60\%$ while the
$21\%$ that never finished becomes in-language error, $18\%\rightarrow39\%$. We do not claim
those failures became successes---under $J$ they are failures either way, which is why $J$
does not move---but they differ in cost. A loop consumes the whole 24,576-token budget and
returns nothing; a terminated wrong answer is short, can be checked or retried, and is what
makes the efficiency gain possible. The counterweight is that a confident wrong answer may
mislead more easily than an obvious non-answer, and no reward geometry fixes that.

\subsection{End-to-end delivery and state distributions}

\begin{table}[!htb]
\centering
\small
\begin{tabular}{@{}lccccccc@{}}
\toprule
 & English & Chinese & Spanish & French & Arabic & Russian & non-English\\
\midrule
Qwen3-4B & $92.9\%$ & $91.4\%$ & $1.4\%$ & $0.0\%$ & $0.0\%$ & $85.7\%$ & $\mathbf{35.7\%}$\\
+ SFT & $94.3\%$ & $85.7\%$ & $87.1\%$ & $87.1\%$ & $80.0\%$ & $82.9\%$ & $\mathbf{84.6\%}$\\
\rowcolor{StageGreen!10}RL: gated & $85.7\%$ & $81.4\%$ & $80.0\%$ & $87.1\%$ & $80.0\%$ & $80.0\%$ & $\mathbf{81.7\%}$\\
\midrule
Qwen3-8B & $92.9\%$ & $91.4\%$ & $0.0\%$ & $0.0\%$ & $0.0\%$ & $61.4\%$ & $\mathbf{30.6\%}$\\
+ SFT (8B) & $94.3\%$ & $85.7\%$ & $88.6\%$ & $84.3\%$ & $85.7\%$ & $87.1\%$ & $\mathbf{86.3\%}$\\
\bottomrule
\end{tabular}

\caption{$J@16$ through the main lineage. The upper block is Qwen3-4B (released, $+$
six-language SFT, $+$ correctness-gated RL) and the lower block Qwen3-8B (released, $+$ SFT);
no RL was run at 8B. The non-English column averages the five primary
non-English languages.}
\label{tab:summary}
\end{table}

\begin{table}[!htb]
\centering
\small
\begin{tabular}{@{}lccp{3.1cm}@{}}
\toprule
diagnostic & additive endpoint & gated continuation & scope \\
\midrule
task--language component & $.7C_{\rm tr}{+}.3\ell_{\rm tr}$ & $C_{\rm tr}(.7{+}.3\ell_{\rm tr})$ & training configuration \\
positive reward mass on wrong & $63.1\%$ & $0.0\%$ & recorded rollouts \\
correct and target-language & $20.5\%$ & $43.8\%$ & late training rollouts \\
mean response tokens & 1960 & 1983 & late training rollouts \\
AIME accuracy & $5.0\%$ & $17.2\%$ & monitored development \\
\bottomrule
\end{tabular}

\caption{Reward-accounting diagnostics from recorded PPO rollouts of the two Qwen3-4B
correctness--language phases---the additive phase, and the gated phase continuing it from
step 200---of one procedure initialized from the Qwen3-4B multilingual SFT checkpoint. These values describe
training dynamics, not held-out capability.}
\label{tab:rltraj}
\end{table}

Table~\ref{tab:summary} tracks $J@16$ through the lineage and Table~\ref{tab:rltraj}
reports the reward-accounting diagnostics behind the RL section.

\paragraph{Difficulty is preserved across the translated prompts.}
Translation could in principle make an item easier or harder rather than merely restate it.
Table~\ref{tab:difficulty} correlates problem-level correctness of the released Qwen3-4B
across all fifteen language pairs of the 70-problem evaluation set: Pearson $r$ ranges from
$0.819$ to $0.988$ with a mean of $0.907$, so the same problems are hard in every language.
The weakest pairs all involve Russian, which is also the language with the largest encoding
factor.

\begin{table}[!htb]
\centering
\small
\begin{tabular}{@{}lcccccc@{}}
\toprule
 & en & zh & es & fr & ar & ru\\
\midrule
en & --- & $.92$ & $.97$ & $.97$ & $.93$ & $.86$\\
zh & $.92$ & --- & $.90$ & $.92$ & $.87$ & $.87$\\
es & $.97$ & $.90$ & --- & $.99$ & $.91$ & $.88$\\
fr & $.97$ & $.92$ & $.99$ & --- & $.91$ & $.88$\\
ar & $.93$ & $.87$ & $.91$ & $.91$ & --- & $.82$\\
ru & $.86$ & $.87$ & $.88$ & $.88$ & $.82$ & ---\\
\bottomrule
\end{tabular}

\caption{Pearson correlation of problem-level correctness between language versions of the
same 70 items, measured on the released Qwen3-4B. High values mean the translation preserved
which problems are difficult.}
\label{tab:difficulty}
\end{table}

\paragraph{Overlap between the RL corpus and the evaluation set.}
Fourteen of the 70 evaluation problems also appear in the 16,754-prompt RL corpus. The
overlap is concentrated in the older half of the evaluation set: 13 of the 40 AMC-23
problems, against 1 of the 30 AIME-2026 problems. The supervised parent never saw that
corpus, so its own drop when those problems are removed measures how much easier they are
for every endpoint, and the RL endpoints must be read against it
(Table~\ref{tab:cleansubset}). Restricting to the clean 56 lowers $J@16$ by $2.4$ points at
the SFT parent and by $0.4$--$3.5$ points at the RL endpoints. The largest excess over the
parent is $1.1$ points, less than half the benchmark effect itself, and the three RL arms
share the same overlapping items, so it cannot account for the differences between them.

\begin{table}[!htb]
\centering
\small
\begin{tabular}{@{}lccccc@{}}
\toprule
endpoint & RL corpus & $J@16$ (70) & $J@16$ (56) & change & excess over SFT\\
\midrule
Qwen3-4B & no & $35.7\%$ & $34.3\%$ & $-1.4\%$ & ---\\
+ SFT & no & $84.6\%$ & $82.1\%$ & $-2.4\%$ & ---\\
RL: correctness only & yes & $15.4\%$ & $15.0\%$ & $-0.4\%$ & $+2.0\%$\\
RL: additive & yes & $83.4\%$ & $80.4\%$ & $-3.1\%$ & $-0.6\%$\\
RL: gated & yes & $81.7\%$ & $78.2\%$ & $-3.5\%$ & $-1.1\%$\\
\bottomrule
\end{tabular}

\caption{$J@16$ on all 70 evaluation problems and on the 56 that do not appear in the RL
training corpus, averaged over the five primary non-English languages. ``Excess over SFT''
is the endpoint's signed change minus the supervised parent's signed change; a negative
value would indicate a memorization advantage.}
\label{tab:cleansubset}
\end{table}

\paragraph{Complete state distributions.}
Table~\ref{tab:ceilings} lists $J@16$ for every endpoint beside the correctness it does not
deliver. The accompanying artifact reports every sample in a disjoint state ledger: delivered,
correct in another language, correct but unfinished, in-language error, in-language
non-termination, English non-termination, English error, or other failure. The released
tables retain both per-language rows and endpoint aggregates, so the compact summaries here
can be reproduced without relying on rounded values.

\begin{table}[!htb]
\centering
\small
\setlength{\tabcolsep}{5pt}
\begin{tabular}{@{}lcccccc@{}}
\toprule
endpoint & en & zh & es & fr & ar & ru \\
\midrule
\texttt{Qwen3-4B} & $92.9\%$\,($0.0\%$) & $91.4\%$\,($0.0\%$) & $1.4\%$\,($90.0\%$) & $0.0\%$\,($92.9\%$) & $0.0\%$\,($91.4\%$) & $85.7\%$\,($0.0\%$) \\
\texttt{Qwen3-8B} & $92.9\%$\,($0.0\%$) & $91.4\%$\,($0.0\%$) & $0.0\%$\,($94.3\%$) & $0.0\%$\,($92.9\%$) & $0.0\%$\,($92.9\%$) & $61.4\%$\,($30.0\%$) \\
\texttt{multilingual SFT} & $94.3\%$\,($0.0\%$) & $85.7\%$\,($0.0\%$) & $87.1\%$\,($0.0\%$) & $87.1\%$\,($0.0\%$) & $80.0\%$\,($1.4\%$) & $82.9\%$\,($1.4\%$) \\
\texttt{multilingual SFT (8B)} & $94.3\%$\,($0.0\%$) & $85.7\%$\,($0.0\%$) & $88.6\%$\,($0.0\%$) & $84.3\%$\,($0.0\%$) & $85.7\%$\,($0.0\%$) & $87.1\%$\,($2.9\%$) \\
\texttt{English-only SFT} & $90.0\%$\,($0.0\%$) & $84.3\%$\,($0.0\%$) & $0.0\%$\,($85.7\%$) & $0.0\%$\,($87.1\%$) & $0.0\%$\,($85.7\%$) & $78.6\%$\,($0.0\%$) \\
\texttt{specialist zh} & $88.6\%$\,($0.0\%$) & $84.3\%$\,($0.0\%$) & -- & -- & -- & -- \\
\texttt{specialist es} & $88.6\%$\,($0.0\%$) & -- & $88.6\%$\,($0.0\%$) & -- & -- & -- \\
\texttt{specialist fr} & $91.4\%$\,($0.0\%$) & -- & -- & $84.3\%$\,($0.0\%$) & -- & -- \\
\texttt{specialist ar} & $90.0\%$\,($0.0\%$) & -- & -- & -- & $82.9\%$\,($0.0\%$) & -- \\
\texttt{specialist ru} & $88.6\%$\,($0.0\%$) & -- & -- & -- & -- & $84.3\%$\,($0.0\%$) \\
\texttt{RL: correctness only} & $81.4\%$\,($0.0\%$) & $70.0\%$\,($12.9\%$) & $0.0\%$\,($81.4\%$) & $0.0\%$\,($82.9\%$) & $0.0\%$\,($87.1\%$) & $7.1\%$\,($77.1\%$) \\
\texttt{RL: additive} & $85.7\%$\,($0.0\%$) & $82.9\%$\,($0.0\%$) & $85.7\%$\,($0.0\%$) & $85.7\%$\,($0.0\%$) & $80.0\%$\,($0.0\%$) & $82.9\%$\,($0.0\%$) \\
\texttt{RL: gated} & $85.7\%$\,($0.0\%$) & $81.4\%$\,($0.0\%$) & $80.0\%$\,($0.0\%$) & $87.1\%$\,($0.0\%$) & $80.0\%$\,($0.0\%$) & $80.0\%$\,($0.0\%$) \\
\bottomrule
\end{tabular}

\caption{$J@16$ for all endpoints, with $C@16-J@16$ in parentheses. Every endpoint is a
Qwen3-4B derivative except the two rows named Qwen3-8B and multilingual SFT (8B);
``multilingual SFT'' is the six-language Qwen3-4B checkpoint from which the correctness-only
and additive runs start. The additive and
gated rows complete the endpoint ledger and are not a matched reward-form comparison.}
\label{tab:ceilings}
\end{table}

\subsection{Efficiency}

\begin{figure}[!htb]
\centering
\includegraphics[width=\textwidth]{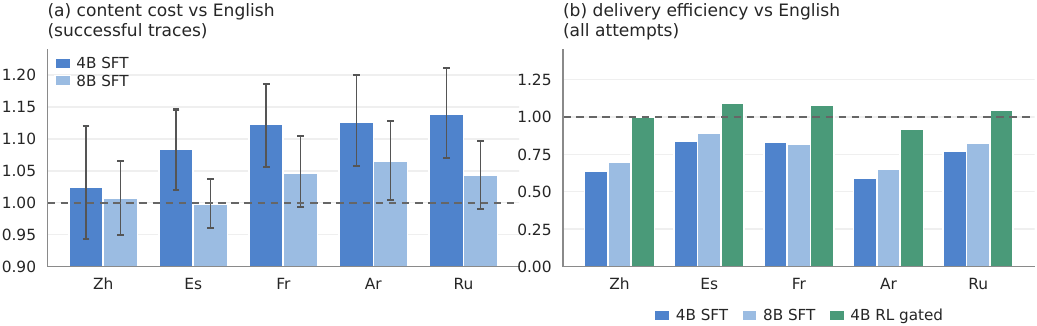}
\caption{Encoding-adjusted content cost among problem-matched delivered traces (a) and
all-attempt delivery efficiency (b), both relative to English at the same endpoint. Bars are
the Qwen3-4B multilingual SFT endpoint, the Qwen3-8B multilingual SFT endpoint, and the
Qwen3-4B correctness-gated RL endpoint. Content is close to parity at
8B, while end-to-end efficiency tracks failed delivery and reaches near parity at the
correctness-gated endpoint. The main paper's efficiency table gives the same quantities without
the encoding adjustment and conditioned on delivered traces.}
\label{fig:efficiency}
\end{figure}

We report both the content cost of a delivered answer and delivery efficiency, which counts
all attempted tokens (Figure~\ref{fig:efficiency}). On the Qwen3-4B SFT endpoint,
encoding-adjusted content cost is only
1.02--1.14 times English across the five target languages; on Qwen3-8B it is near parity. Delivery
efficiency remains lower where failure rates are higher, and improves after RL reduces
non-termination. Thus successful target-language traces are not the primary source of the
initial access gap; failed attempts are part of the end-to-end cost.

\paragraph{Every efficiency number in four accounting conventions.}
Two choices are made when tokens are counted, and both change the number. The first is
whether to divide out the per-language encoding factor: adjusted tokens answer whether the
model produced more reasoning, whereas raw tokens are what a user is billed for. The second
is whether to charge the tokens of attempts that were never delivered: the all-attempt
convention charges them, the $J$-only convention does not. The main paper reports all four for every stage, and Table~\ref{tab:efflang} resolves
them by language. The unit is
the same throughout---joint successes per 1,000 tokens, expressed relative to the same
endpoint's English---so the conventions are directly comparable, and they are linked exactly
by
\[
\underbrace{\mathrm{eff}_{\text{all}}}_{\text{end-to-end}}
=
\underbrace{\mathrm{eff}_{J}}_{\text{cost of a delivered solution}}
\times
\underbrace{\mathrm{yield}}_{\text{tokens inside delivered traces}/\text{tokens generated}} .
\]
Three facts follow. First, the parity claimed at the gated endpoint is specific to adjusted
tokens: in raw tokens the same endpoint reaches $87\%$ of English, and the residual is
exactly the per-language encoding factor. Second, conditioning on delivery removes the
penalty entirely: a delivered non-English solution costs $117\%$ of English efficiency in
adjusted tokens and $100\%$ in raw tokens at the gated endpoint, and the Qwen3-8B SFT
endpoint reproduces this at $111\%$ and $94\%$. Third, the end-to-end gap is therefore
carried by yield rather than by length. Table~\ref{tab:yield} gives yield per language: at
the released Qwen3-4B checkpoint, Spanish, French, and Arabic convert $0.0\%$--$0.1\%$ of
their generated tokens into a delivered solution, against $59.7\%$ in English, and after SFT
the lowest-yield languages are the ones that loop most.

The same distinction applies to content cost, which the main text reports in adjusted tokens.
In raw tokens, the paired content cost of a delivered solution at the Qwen3-4B SFT endpoint
is 1.03--1.50 times English, which is the tokenizer-fertility effect that prior work already
documents; the encoding-adjusted values in the same rows are 1.02--1.14. We therefore make
the convention explicit wherever the number appears rather than presenting either as the
efficiency of the model. Full per-language values in both conventions, including intervals,
are released with \texttt{analysis/A11\_efficiency\_split} (adjusted content and all-attempt
delivery) and \texttt{analysis/A29\_token\_efficiency} (all four conventions, yield, and raw
content cost).

\begin{table}[!htb]
\centering
\small
\begin{tabular}{@{}llccccc@{}}
\toprule
stage & tokens charged & zh & es & fr & ar & ru\\
\midrule
Qwen3-4B & all attempts & $109$ (108) & $0$ (0) & $0$ (0) & $0$ (0) & $128$ (98)\\
 & $J$-only & $131$ (131) & $\dagger$ & -- & -- & $169$ (128)\\
+ SFT & all attempts & $64$ (63) & $84$ (71) & $83$ (69) & $59$ (48) & $77$ (59)\\
 & $J$-only & $121$ (120) & $104$ (88) & $103$ (86) & $113$ (92) & $108$ (82)\\
RL: correctness only & all attempts & $44$ (44) & $0$ (0) & $0$ (0) & $0$ (0) & $1$ (1)\\
 & $J$-only & $134$ (134) & -- & -- & -- & $\dagger$\\
RL: additive & all attempts & $101$ (101) & $102$ (86) & $99$ (83) & $94$ (76) & $101$ (77)\\
 & $J$-only & $115$ (114) & $103$ (87) & $102$ (86) & $102$ (83) & $101$ (77)\\
RL: gated & all attempts & $100$ (99) & $109$ (92) & $108$ (90) & $91$ (74) & $105$ (80)\\
 & $J$-only & $123$ (122) & $118$ (100) & $114$ (95) & $117$ (96) & $113$ (86)\\
\addlinespace[2pt]
Qwen3-8B & all attempts & $99$ (99) & $0$ (0) & $0$ (0) & $0$ (0) & $35$ (32)\\
 & $J$-only & $120$ (120) & -- & -- & -- & $163$ (124)\\
+ SFT (8B) & all attempts & $70$ (70) & $89$ (75) & $82$ (69) & $65$ (53) & $82$ (62)\\
 & $J$-only & $112$ (112) & $110$ (93) & $109$ (91) & $111$ (91) & $110$ (84)\\
\bottomrule
\end{tabular}

\caption{Delivery efficiency by language, as a percentage of the same endpoint's English
efficiency. Each cell is encoding-adjusted, with the raw-token value in parentheses.
$\dagger$ marks a $J$-only cell with fewer than 20 delivered samples out of 1,120, which is
too thin to report; \textit{--} marks zero delivered samples. Chinese is nearly unaffected by
the adjustment because its encoding factor is $1.003$, whereas Russian is affected most at
$1.314$.}
\label{tab:efflang}
\end{table}

\begin{table}[!htb]
\centering
\small
\begin{tabular}{@{}lcccccc@{}}
\toprule
stage & en & zh & es & fr & ar & ru\\
\midrule
Qwen3-4B & $59.7$ & $49.4$ & $0.1$ & $0.0$ & $0.0$ & $45.7$\\
+ SFT & $45.9$ & $24.2$ & $37.0$ & $37.0$ & $24.0$ & $32.9$\\
RL: correctness only & $37.1$ & $12.1$ & $0.0$ & $0.0$ & $0.0$ & $0.5$\\
RL: additive & $35.6$ & $31.4$ & $35.4$ & $34.3$ & $32.8$ & $35.4$\\
RL: gated & $39.8$ & $32.4$ & $36.7$ & $37.6$ & $31.0$ & $36.8$\\
\addlinespace[2pt]
Qwen3-8B & $61.5$ & $50.8$ & $0.0$ & $0.0$ & $0.0$ & $15.6$\\
+ SFT (8B) & $51.6$ & $32.2$ & $41.8$ & $38.8$ & $30.2$ & $38.4$\\
\bottomrule
\end{tabular}

\caption{Token yield: the percentage of generated tokens that sit inside a delivered
solution, by language and stage, in raw tokens. This quantity is dimensionless and
tokenizer-independent, and it is the factor that separates the two conventions in the
main paper's efficiency table.}
\label{tab:yield}
\end{table}

\section{Prompts and a Trace Case Study}\label{app:prompts}

\subsection{Forced-target-language evaluation prompt}

Every language uses the same three-part template, translated in full rather than appending
an English instruction to a localized problem. The literal token \texttt{Answer:} is kept
constant so that final-answer extraction does not depend on language. The model's standard
chat template is then applied with thinking enabled.

\medskip
\noindent\fcolorbox{StageGray}{StageGray!6}{%
\begin{minipage}{.94\textwidth}
\textbf{Localized instruction.} Solve the following mathematics problem step by step,
reasoning entirely in \textsc{language}. The last line of your response must have the form
\texttt{Answer: \$Answer}, where \texttt{\$Answer} is the answer to the problem.

\medskip
\textbf{Localized problem.} \texttt{\{parallel problem statement\}}

\medskip
\textbf{Localized suffix.} Remember to put your answer on its own line after
\texttt{Answer:}.
\end{minipage}}

\subsection{The same problem in six languages}\label{app:cases}

\begin{figure*}[t]
\centering
\includegraphics[width=\textwidth]{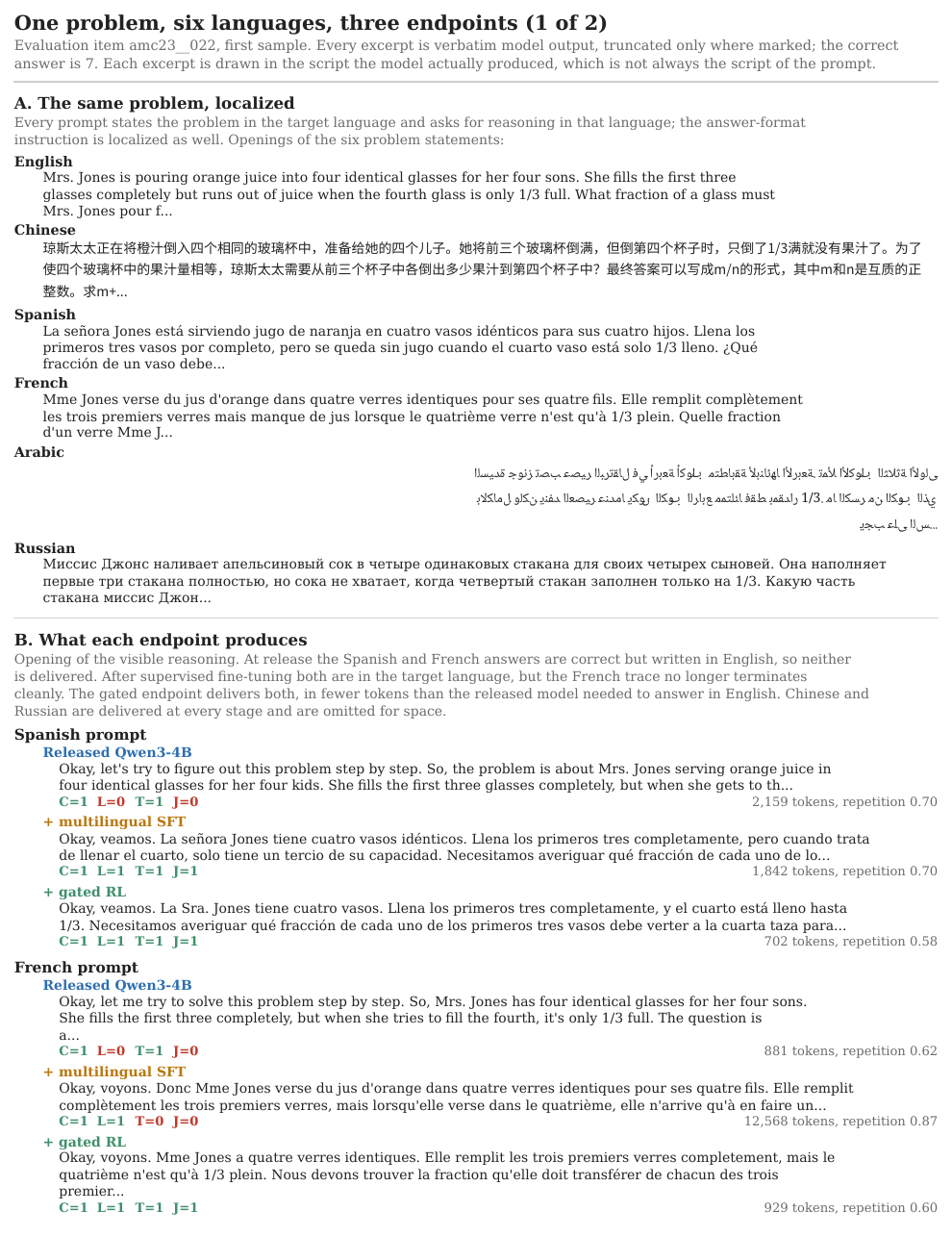}
\caption{One evaluation item (\texttt{amc23\_\_022}) through the lineage: the localized
problem in six languages, and the Spanish and French traces at three endpoints. Chinese
and Arabic are drawn with Noto fonts inside the figure because the manuscript is compiled
with pdflatex, which has no font for those scripts on our system; the script each excerpt
appears in is the script the model actually produced.}
\label{fig:casesa}
\end{figure*}

\begin{figure*}[t]
\centering
\includegraphics[width=\textwidth]{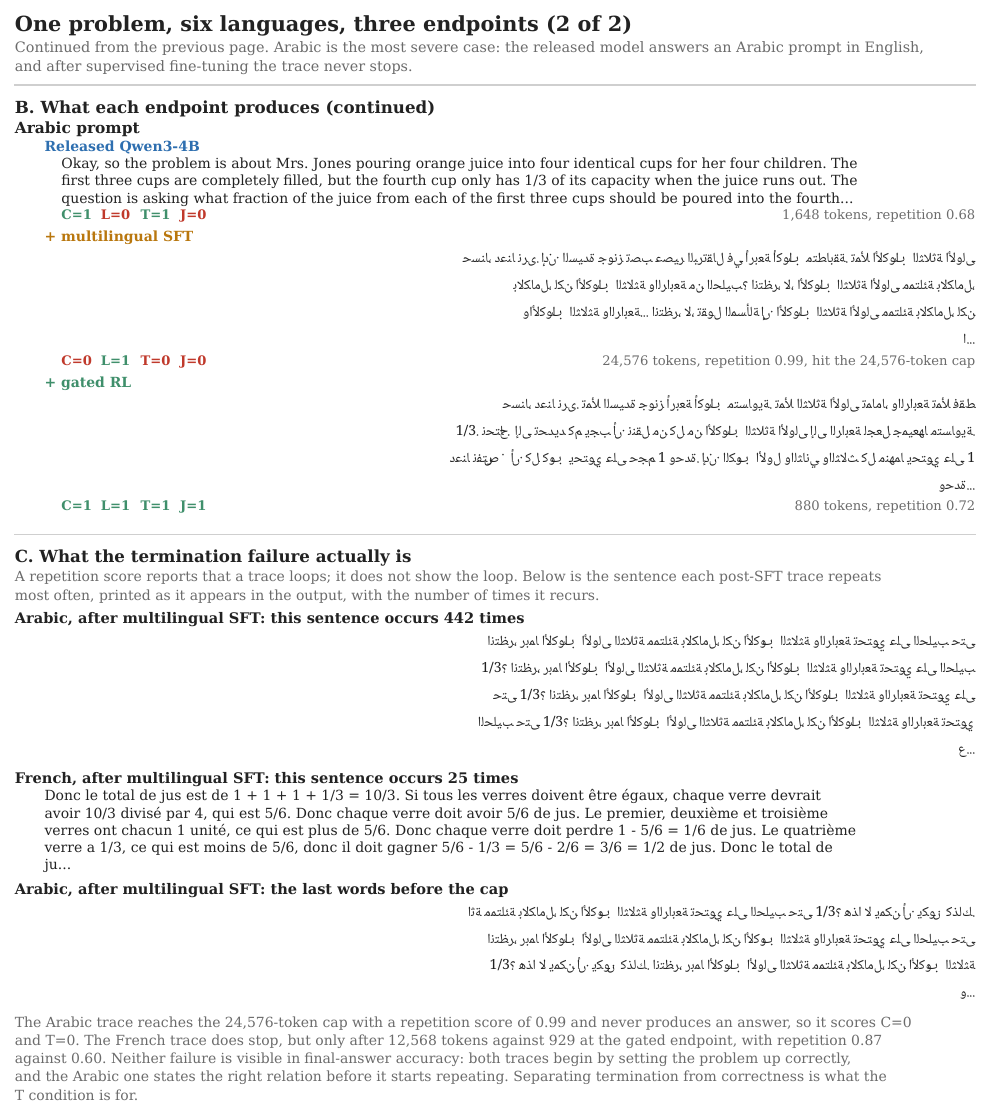}
\caption{The same item continued: the Arabic traces at three endpoints, and the literal
content of the post-SFT termination failure with the number of times each repeated
sentence recurs.}
\label{fig:casesb}
\end{figure*}

Figures~\ref{fig:casesa} and~\ref{fig:casesb} show what the readouts correspond to in
text. They take a single evaluation item, give the localized problem in all six languages,
and print the opening of the visible reasoning produced by the released checkpoint, the
supervised checkpoint, and the gated endpoint for the three languages where the released
model pivots to English. The excerpts are verbatim, truncated only for length.

The item was chosen because every phenomenon the paper measures occurs on it at once. At
release, the Spanish, French, and Arabic answers are correct and written in English, so
$C=1$ but $L=0$ and nothing is delivered; the Chinese and Russian answers are already
delivered. After supervised fine-tuning, all five are written in the target language, but
the French trace takes 12,568 tokens and is scored as looping, and the Arabic trace runs to
the 24,576-token cap with a repetition score of $99\%$ and never produces an answer. The
gated endpoint delivers all five in 536--929 tokens, fewer on average than the released
model used to answer these items in English. A reader who only saw final-answer accuracy would count the release row
as a success and the Arabic supervised row as an ordinary wrong answer.

The second page prints the loop itself rather than a score for it. In the Arabic
supervised trace, one sentence---a restatement of the problem setup, phrased as a
question---occurs 442 times; in the French trace the repeated sentence occurs 25 times.
Both traces begin by setting the problem up correctly, which is why this failure is
invisible to a metric that reads only the final answer.

These two pages cannot establish a rate, and this item is not a random draw: it was
selected to show the transitions together. The statistics behind each transition are in
the main results.

\end{document}